\documentclass[10pt,twocolumn,letterpaper]{article}
\usepackage[pagenumbers]{cvpr} % To force page numbers, e.g. for an arXiv version

\usepackage[accsupp]{axessibility}
\usepackage[]{multibib}
\usepackage{tocloft}
\usepackage{multibib}
\usepackage{color, colortbl}
\usepackage[dvipsnames]{xcolor}
\newcommand{\red}[1]{{\color{red}#1}}
\newcommand{\blue}[1]{{\color{blue}#1}}

\newcommand{\methodname}{SiTH}
\definecolor{cvprblue}{rgb}{0.21,0.49,0.74}
\usepackage[pagebackref,breaklinks,colorlinks,citecolor=cvprblue]{hyperref}
\def\paperID{***} % *** Enter the Paper ID here
\def\confName{CVPR}
\def\confYear{2024}

\title{SiTH: Single-view Textured Human Reconstruction \\
with Image-Conditioned Diffusion}

\author{Hsuan-I Ho~~~
\and
Jie Song~~~
\and
Otmar Hilliges~~~
    \vspace{0.1em}
\and
~~~~~~~~Department of Computer Science, ETH Zürich~~~~~~~~
\\
{\small\url{https://ait.ethz.ch/sith}}
}

\begin{document}
\twocolumn[{%
\renewcommand\twocolumn[1][]{#1}%
\maketitle
\begin{center}
    \vspace{-1.5em}
    \centering
    \captionsetup{type=figure}
    \includegraphics[width=0.95\textwidth]{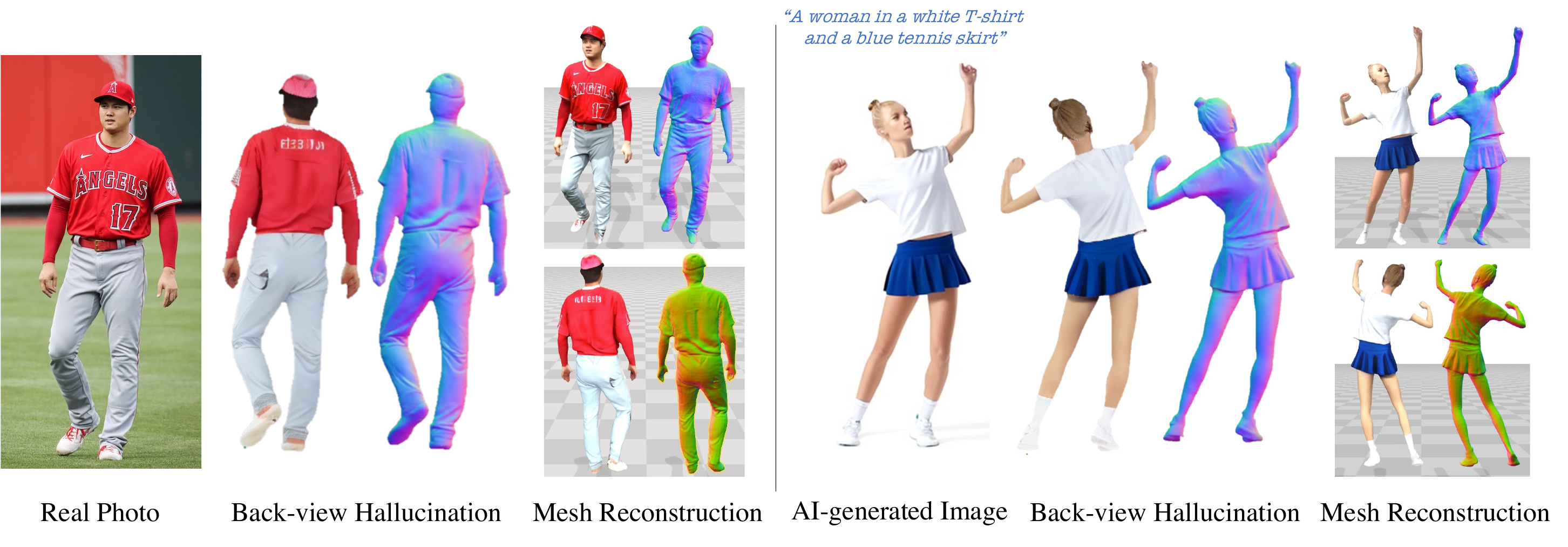}
    \vspace{-1em}
    \caption{\textbf{Single-view textured human reconstruction}.~\methodname ~is a novel pipeline for creating high-quality and fully textured 3D human meshes from single images. We first hallucinate back-view appearances through an image-conditioned diffusion model, followed by the reconstruction of full-body textured meshes using both the front and back-view images. Our pipeline enables the creation of lifelike and diverse 3D humans from unseen photos (\emph{left}) and AI-generated images (\emph{right}).}
    \label{fig:teaser}
\end{center}%
}]

\addtocontents{toc}{\protect\setcounter{tocdepth}{0}}
\begin{abstract}
A long-standing goal of 3D human reconstruction is to create lifelike and fully detailed 3D humans from single-view images.
The main challenge lies in inferring unknown body shapes, appearances, and clothing details in areas not visible in the images. 
To address this, we propose~\methodname, a novel pipeline that uniquely integrates an image-conditioned diffusion model into a 3D mesh reconstruction workflow.
At the core of our method lies the decomposition of the challenging single-view reconstruction problem into generative hallucination and reconstruction subproblems.
For the former, we employ a powerful generative diffusion model to hallucinate unseen back-view appearance based on the input images.
For the latter, we leverage skinned body meshes as guidance to recover full-body texture meshes from the input and back-view images.
~\methodname~requires as few as 500 3D human scans for training while maintaining its generality and robustness to diverse images.
Extensive evaluations on two 3D human benchmarks, including our newly created one, highlighted our method's superior accuracy and perceptual quality in 3D textured human reconstruction.
\end{abstract}
    
\vspace{-1em}
\section{Introduction}
\label{sec:intro}
With the growing popularity of 3D and virtual reality applications, there has been increasing interest in creating realistic 3D human models. In general, crafting 3D humans is labor-intensive, time-consuming, and requires collaboration from highly skilled professionals. To bring lifelike 3D humans to reality and to support both expert and amateur creators in this task, it is essential to enable users to create textured 3D humans from simple 2D images or photos.

Reconstructing a fully textured human mesh from a single-view image presents an ill-posed problem with two major challenges. Firstly, the appearance information required for generating texture in unobserved regions is missing. Secondly, 3D information for mesh reconstruction, such as depth, surface, and body pose, becomes ambiguous in a 2D image.
Previous efforts~\cite{saito2019pifu,zheng2021pamir,alldieck2022phorhum} attempted to tackle these challenges in a data-driven manner, focusing on training neural networks with image-mesh pairs. 
However, these approaches struggle with images featuring unseen appearances or poses, due to limited 3D human training data.
More recent studies~\cite{saito2020pifuhd,zheng2021pamir,xiu2022icon} introduced additional 3D reasoning modules to enhance robustness against unseen poses. Yet, generating realistic and full-body textures from unseen appearances still remains an unsolved problem. 

To address the above challenges, we propose~\methodname, a novel pipeline that integrates an image-conditioned diffusion model to reconstruct lifelike 3D textured humans from monocular images.
At the core of our approach is the decomposition of the challenging single-view problem into two subproblems: generative back-view hallucination and mesh reconstruction. This decomposition enables us to exploit the generative capability of pretrained diffusion models to guide full-body mesh and texture reconstruction. The workflow is depicted in~\cref{fig:teaser}. Given a front-view image, the first stage involves hallucinating a perceptually consistent back-view image using image-conditioned diffusion. The second stage reconstructs full-body mesh and texture, utilizing both the front and back-view images as guidance. 

More specifically, we employ the generative capabilities of pretrained diffusion models (e.g. Stable Diffusion~\cite{rombach2022high}) to infer unobserved back-view appearances for full-body 3D reconstruction.
The primary challenge in ensuring the realism of 3D meshes lies in generating images that depict spatially aligned body shapes and perceptually consistent appearances with the input images.
While diffusion models demonstrate impressive generative abilities with text conditioning, they are limited in producing desired back-view images using the frontal images as \emph{image conditions}. 
To overcome this, we adapt the network architecture to enable conditioning on frontal images and introduce additional trainable components following ControlNet~\cite{zhang2023adding} to provide pose and mask control. To fully tailor this model to our task while retaining its original generative power, we carefully fine-tune the diffusion model using multi-view images rendered from 3D human scans. Complementing this generative model, we develop a mesh reconstruction module to recover full-body textured mesh from front and back-view images. We follow prior work in handling 3D ambiguity through normal~\cite{saito2020pifuhd} and skinned body~\cite{xiu2022icon,zheng2021pamir} guidance. It is worth noting that the models for both subproblems are trained using the same public THuman2.0~\cite{tao2021function4d} dataset, which consists of as few as 500 scans.

To advance research in single-view human reconstruction, we created a new benchmark based on the high-quality CustomHumans~\cite{ho2023custom} dataset and conducted comprehensive evaluations against state-of-the-art methods. Compared to existing end-to-end methods~\cite{saito2019pifu,zheng2021pamir,alldieck2022phorhum}, our two-stage pipeline can recover full-body textured meshes, including back-view details, and demonstrates robustness to unseen images. In contrast to time-intensive diffusion-based optimization methods~\cite{liu2023zero,kim2023chupa,huang2024tech}, our pipeline efficiently produces high-quality textured meshes in under two minutes. 
Moreover, we explored applications combining text-guided diffusion models, showing~\methodname's versatility in 3D human creation. Our contributions are summarized as follows:

\begin{itemize}
    \item We introduce~\methodname, a single-view human reconstruction pipeline capable of producing high-quality, fully textured 3D human meshes within two minutes.
    \item Through decomposing the single-view reconstruction task, \methodname~can be efficiently trained with public 3D human scans and is more robust to unseen images.
    \item We establish a new benchmark featuring more diverse subjects for evaluating textured human reconstruction.
\end{itemize}

\begin{figure*}[t]
\centering
\includegraphics[width=\linewidth]{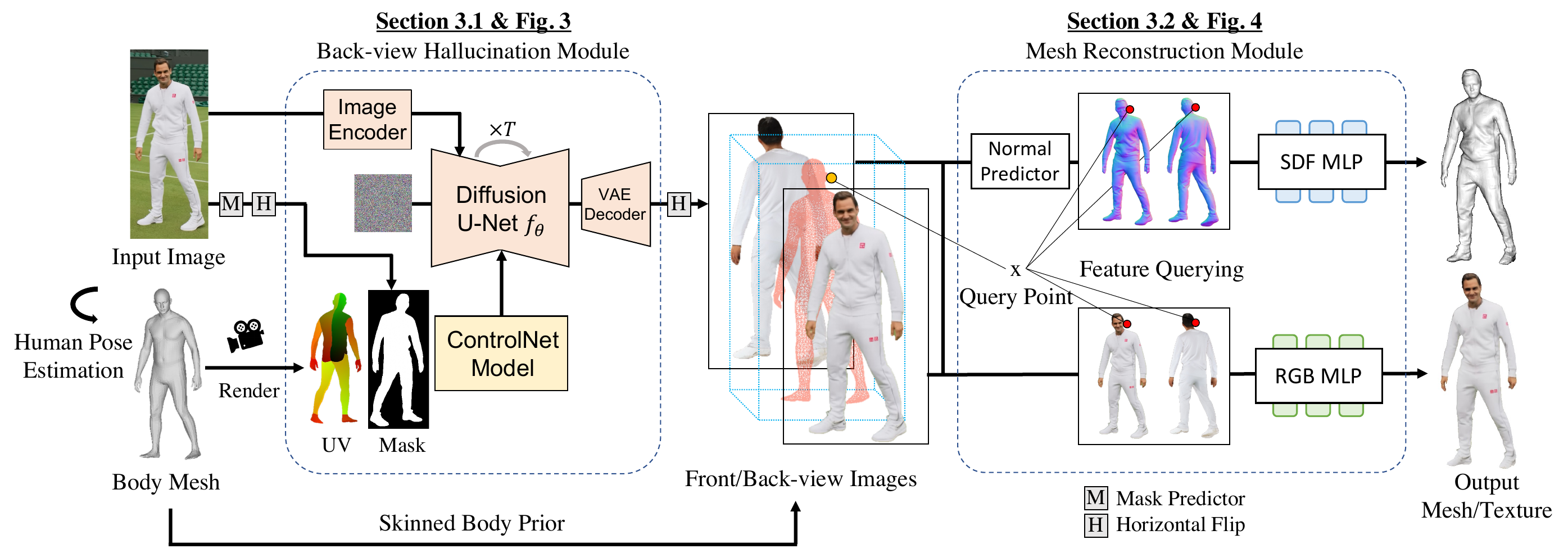}
    \vspace{-2em}
\caption{\textbf{Method overview}.~\methodname~is a two-stage pipeline composed of back-view hallucination and mesh reconstruction. The back-view hallucination module samples perceptually consistent back-view images through an iterative denoising process conditioned on the input image, UV map, and silhouette mask (\cref{sec:diffusion}). Based on the input and generated back-view images, the mesh reconstruction module recovers a full-body mesh and textures leveraging a skinned body prior as guidance (\cref{sec:mesh}). Note that both modules in the pipeline can be trained with the same public 3D human dataset and generalize unseen images.
}
\label{fig:framework}
\vspace{-1.5em}
\end{figure*}
%\vspace{-0.5em}
\section{Related Work}
\label{sec:work}
%-------------------------------------------------------------------------
\vspace{-.5em}
\paragraph{Single-view human mesh reconstruction.} 
Reconstructing 3D humans from monocular inputs~\cite{tian2023recovering, weng_humannerf_2022_cvpr,jiang2022selfrecon,dong2022pina,guo2023vid2avatar,shen2023xavatar,jiang2022instantavatar,jiang2022neuman} has gained more popularity in research. In this context, we focus on methods that recover 3D human shapes, garments, and textures from a single image.
As a seminal work, Saito~\etal~\cite{saito2019pifu} first proposed a data-driven method with pixel-aligned features and neural fields~\cite{xie2022neural}. Its follow-up work PIFuHD~\cite{saito2020pifuhd} further improved this framework with high-res normal guidance. Later approaches extended this framework with additional human body priors. For instance, PaMIR~\cite{zheng2021pamir} and ICON~\cite{xiu2022icon} utilized skinned body models~\cite{SMPL:2015, pavlakos2019expressive} to guide 3D reconstruction. ARCH~\cite{huang2020arch}, ARCH++~\cite{he2021arch++}, and CAR~\cite{liao2023car} transformed global coordinates into the canonical coordinates to allow for reposing. 
PHOHRUM~\cite{alldieck2022phorhum} and S3F~\cite{corona2023s3f} further disentangled shading and albedo to enable relighting. 
Another line of work replaced the neural representations with conventional Poisson surface reconstruction~\cite{kazhdan2006poisson,kazhdan2013screened}. ECON~\cite{xiu2023econ} and 2K2K~\cite{han2023high} trained normal and depth predictors to generate front and back 2.5D point clouds. The human mesh is obtained by fusing these point clouds with body priors and 3D heuristics.
However, none of these methods produce realistic full-body texture and geometry in the unobserved regions. Our pipeline addresses this problem by incorporating a generative diffusion model into the 3D human reconstruction workflow. 
\vspace{-1.5em}
%-------------------------------------------------------------------------
\paragraph{3D generation with 2D diffusion models.}
Diffusion models~\cite{rombach2022high,saharia2022photorealistic,ramesh2022hierarchical,podell2023sdxl} trained with large collections of images have demonstrated unprecedented capability in creating 3D objects from text prompts. Most prior work~\cite{poole2023dreamfusion,wang2023score,lin2023magic3d,Metzer_2023_CVPR,instructnerf2023,Chen_2023_ICCV} followed an optimization workflow to update 3D representations (e.g. NeRF~\cite{mildenhall2020nerf}, SDF tetrahedron~\cite{shen2021dmtet}) via neural rendering~\cite{tewari2022advances} and a score distillation sampling (SDS)~\cite{poole2023dreamfusion} loss. While some methods~\cite{zhang2023avatarverse,cao2023dreamavatar,jiang2023avatarcraft,Svitov_2023_ICCV,albahar2023humansgd,huang2024tech} applied this workflow to human bodies, they cannot produce accurate human bodies and appearances due to the ambiguity of text-conditioning. More recent work~\cite{liu2023zero,qian2023magic123} also tried to extend this workflow with more accurate image-conditioning. However, we show that they struggle to recover human clothing details and require a long optimization time.
Most related to our work is Chupa~\cite{kim2023chupa}, which also decomposes its pipeline into two stages.
Note that Chupa is an optimization-based approach that relies on texts and cannot model colors. We address these issues by introducing an image-conditioning strategy and model.
Most importantly, our method swiftly reconstructs full-texture human meshes without any optimization process.
%-------------------------------------------------------------------------
\vspace{-1.5em}
\paragraph{Diffusion models adaptation.}
Foundation models~\cite{devlin2018bert,brown2020language,he2022masked,kirillov2023segany} trained on large-scale datasets have been shown to be adaptable to various downstream tasks. Following this trend, pretrained diffusion models~\cite{rombach2022high,saharia2022photorealistic,ramesh2022hierarchical,podell2023sdxl} have become common backbones for generative modeling. For instance, they can be customized by finetuning with a small collection of images~\cite{hu2022lora,kumari2022customdiffusion,Ruiz_2023_CVPR}. ControlNet~\cite{zhang2023adding} introduced additional trainable plugins to enable image conditioning such as body skeletons. While these strategies have been widely adopted, none of them directly fit our objective.
More relevant to our task is DreamPose~\cite{dreampose_2023}, which utilizes DensePose~\cite{guler2018densepose} images as conditions to repose input images. However, it cannot handle out-of-distribution images due to overfitting. Similarly, Zero-1-to-3~\cite{liu2023zero} finetunes a diffusion model with multi-view images to allow for viewpoint control. However, we show that viewpoint conditioning is not sufficient for generating consistent human bodies. Our model addresses this issue by providing accurate body pose and mask conditions for back-view hallucination.

\vspace{-0.5em}
\section{Methodology}
\label{sec:method}
\label{sec:overview}
\vspace{-0.5em}
\paragraph{Method overview.}
Given an input image of a human body and estimated SMPL-X~\cite{pavlakos2019expressive} parameters,~\methodname~produces a full-body textured mesh. 
This mesh not only captures the observed appearances but also recovers geometric and textural details in unseen regions, such as clothing wrinkles on the back.
The pipeline is composed of two modules and is summarized in~\cref{fig:framework}. 
In the first stage, we hallucinate unobserved appearances leveraging the generative power of an image-conditioned diffusion model (\cref{sec:diffusion}).
In the second stage, we reconstruct a full-body textured mesh given the input front-view image and the generated back-view image as guidance (\cref{sec:mesh}).
Notably, both modules are efficiently trained with 500 textured human scans in THuman2.0~\cite{tao2021function4d}.

\vspace{-0.5em}
\subsection{Back-view Hallucination}
\label{sec:diffusion}
\vspace{-0.5em}
\paragraph{Preliminaries.}
Given an input front-view image $I^F\in\mathbb{R}^{H\times W \times3}$, our goal is to infer a back-view image $I^B\in\ \mathbb{R}^{H\times W \times3}$ which depicts unobserved body appearances. 
This task is under-constrained since there are multiple possible solutions to the same input images. Taking this perspective into account, we leverage a \textbf{latent diffusion model (LDM)}~\cite{rombach2022high} to learn a conditional distribution of back-view images given a front-view image.
First, a VAE autoencoder, consisting of an encoder $\mathcal{E}$ and a decoder $\mathcal{D}$, is pretrained on a corpus of 2D natural images through image reconstruction, i.e. $\tilde{I} = \mathcal{D}(\mathcal{E}(I))$. 
Afterwards, an LDM learns to produce a latent code $z$ within the VAE latent distribution $z = \mathcal{E}(I)$ from randomly sampled noise.
To sample an image, a latent code $\tilde{z}$ is obtained by iteratively denoising Gaussian noise. The final image is reconstructed through the decoder, i.e.,  $ \tilde{I}=\mathcal{D}(\tilde{z})$.

\begin{figure}[t]
\centering
\includegraphics[width=\linewidth]{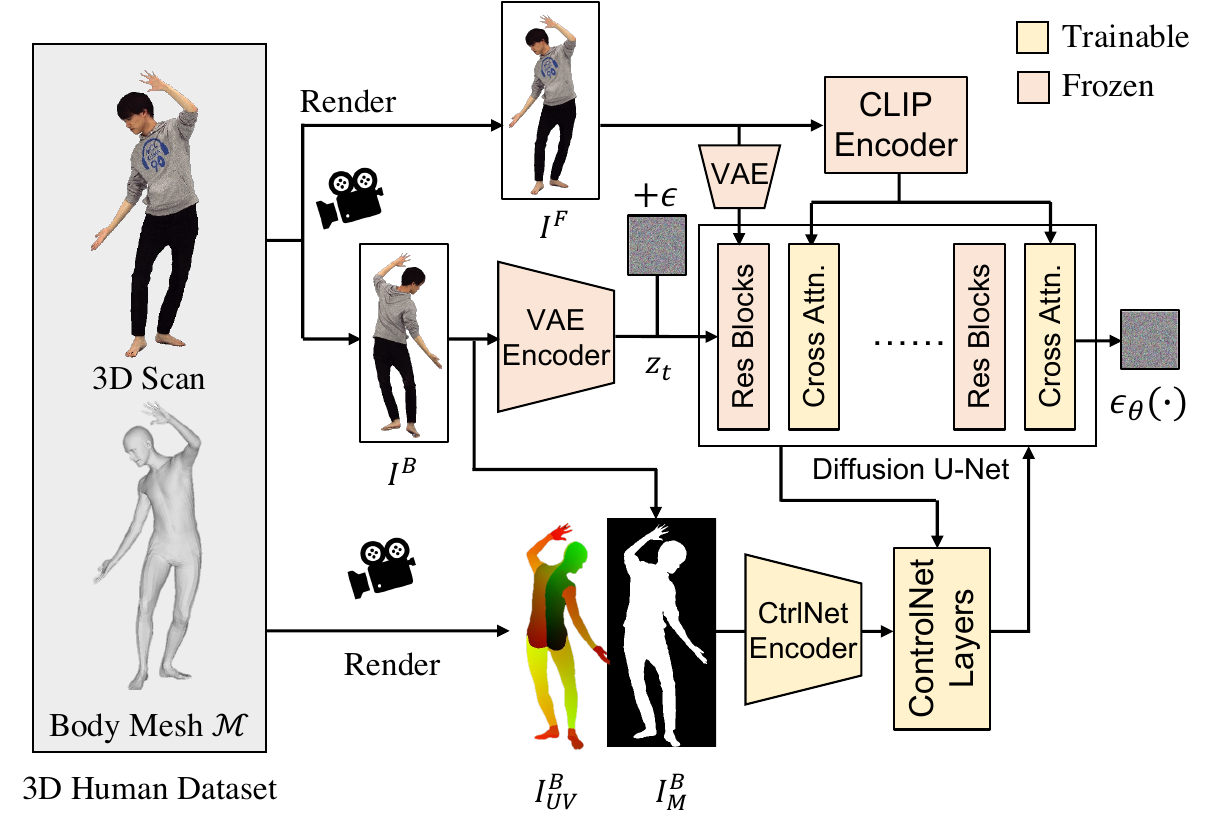}
    \vspace{-2em}
\caption{\textbf{Training of back-view hallucination module}.
We employ a pretrained LDM and ControlNet architecture to enable image conditioning. To train our model, we render training pairs of conditional images $I^F$ and ground-truth images $I^B$ from 3D human scans. Given a noisy image latent $z_t$, the model predicts added noise $\epsilon$ given the conditional image $I^F$, UV map $I^B_{UV}$, and mask $I^B_{M}$ as conditions.
We train the ControlNet model and cross-attention layers while keeping other parameters frozen.
}
\label{fig:diffusion}
\vspace{-1.5em}
\end{figure}

\vspace{-1.5em}
\paragraph{Image-conditioned diffusion model.}
Simply applying the LDM architecture to our task is not sufficient since our goal is to learn a conditional distribution of back-view images given an input conditional image.
To this end, we make several adaptations to allow for image-conditioning as shown in~\cref{fig:diffusion}.
First, we utilize the pretrained CLIP~\cite{radford2021learning} image encoder and VAE encoder $\mathcal{E}$ to extract image features from the front-view image (i.e., $I^F$). These image features are used for conditioning the LDM, ensuring the output image shares a consistent appearance with the input image.
Second, we follow the idea of ControlNet~\cite{zhang2023adding} and propose to use a UV map ($I^B_{UV} \in \mathbb{R}^{H\times W \times3}$) and a silhouette mask ($I^B_M \in \mathbb{R}^{H\times W }$) from the back view as additional conditions.
These conditional signals provide additional information that ensures the output image has a similar body shape and pose to the conditional input image.
\vspace{-1.5em}
\paragraph{Learning hallucination from pretraining.}
Another challenge in training an image-conditioned LDM is data. Training the model from scratch is infeasible due to the requirement of a large number of paired images rendered from 3D textured human scans. Inspired by the concept of learning from large-scale pretraining~\cite{devlin2018bert,he2022masked}, we build our image-conditioned LDM on top of a pretrained diffusion U-Net~\cite{rombach2022high}. 
We utilize the finetuning strategy~\cite{kumari2022customdiffusion, zhang2023adding} to optimize cross-attention layers and ControlNet parameters while keeping most of the other parameters frozen (see \cref{fig:diffusion}).
The design and training strategy of our image-conditioned diffusion model enables hallucinating plausible back-view images that are \emph{cosistent} with the frontal inputs. 

\vspace{-2em}
\paragraph{Training and inference.}
To generate pairwise training images from 3D human scans, we sample camera view angles and use orthographic projection to render RGBA images from 3D scans and UV maps from their SMPL-X fits.
Given a pair of images rendered by a frontal and its corresponding back camera, the first image serves as the conditional input $I^F$ while the other one is the ground-truth image $I^B$. 
During training, the ground-truth latent code $z_0 = \mathcal{E}(I^B)$ is perturbed by the diffusion process in t time steps, resulting in a noisy latent $z_t$. The image-conditoned LDM model $\epsilon_\theta$ aims to predict the added noise $\epsilon$ given the noisy latent $z_t$, the time step $t\sim[0,1000]$, the conditional image $I^F$, the silhouette mask $I^B_{M}$, and the UV map $I^B_{UV}$ (See~\cref{fig:diffusion}). The objective function for fine-tuning can be represented as:
\begin{equation}
    \mathop{\min_{\theta}} \mathbb{E}_{z\sim\mathcal{E}(I),t,\epsilon\sim\mathcal{N}(\mathbf{0,I})}  \left \| \epsilon - \epsilon_\theta (z_t, t, I^F, I^B_{UV}, I^B_M) \right \|_{2}^{2}.
\end{equation}
At test time, we obtain $I^B_{UV}, I^B_{M}$ from an off-the-shelf pose predictor~\cite{cai2023smplerx} and segmentation model~\cite{kirillov2023segany}. To infer a back-view image, we sample a latent $\tilde{z}_0$ by performing the iterative denoising process starting from a Gaussian noise $z_T \sim\mathcal{N}(\mathbf{0, I})$. The back-view image can obtained by:
\begin{equation}
    \tilde{I}^B = \mathcal{D}(\tilde{z}_0) = \mathcal{D}( f_\theta(z_T, I^F, I^B_{UV}, I^B_M) ),
\end{equation}
where $f_\theta$ is a function representing the iterative denoising process of our image-conditioned LDM (See \cref{fig:framework} left).
\begin{figure}[t]
\centering
\includegraphics[width=0.8\linewidth]{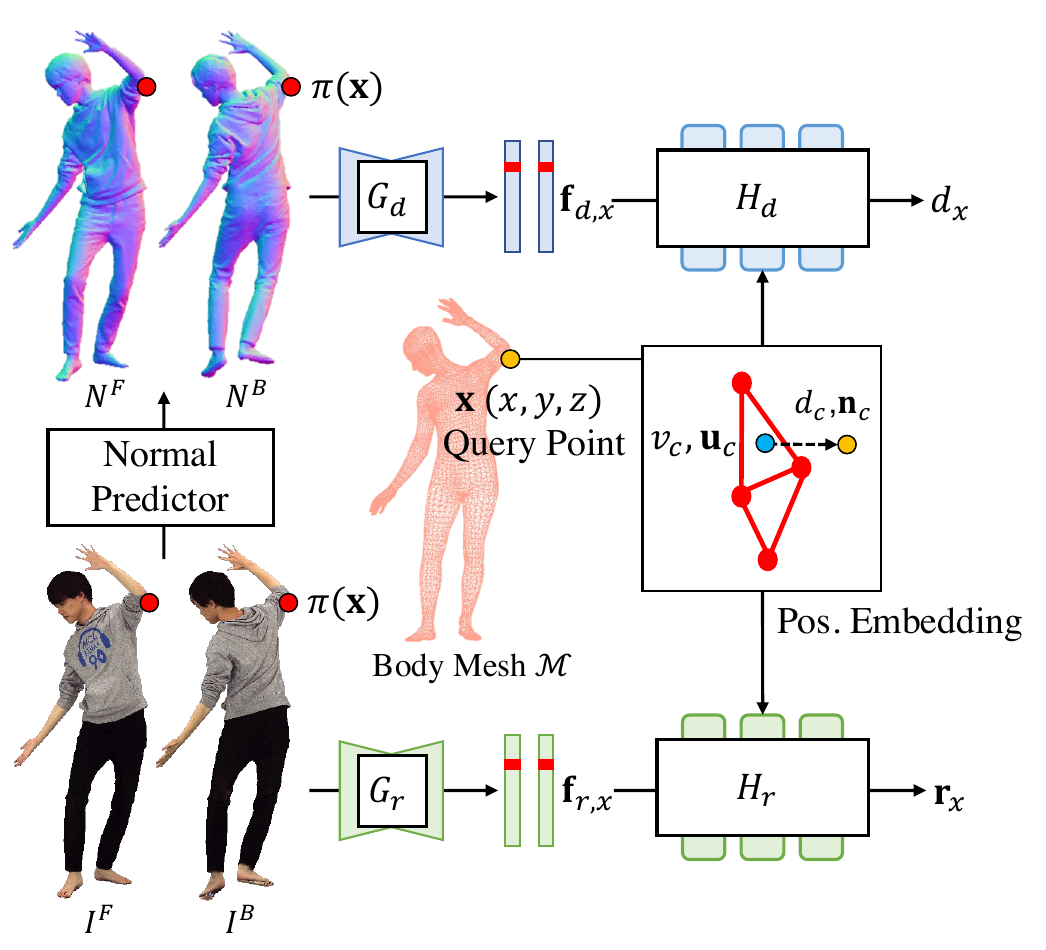}
    \vspace{-1em}
\caption{\textbf{Mesh reconstruction module}. Given front and back-view images ($I^F,I^B$) we predict their normal images ($N^F,N^B$) through a learned normal predictor. A 3D point $\mathbf{x}$ is projected onto these images for querying pixel-aligned features ($\mathbf{f}_{d,x},\mathbf{f}_{r,x}$). To leverage human body mesh as guidance, we embed the point $\mathbf{x}$ into the local UV coordinates $\mathbf{u}_c$, vector $\mathbf{n_c}$, distance $d_c$, and visibility $v_c$. Finally, two decoders ($H_d,H_r$) predict SDF and RGB values at $\mathbf{x}$ given the positional embedding and pixel-aligned features.
}
\label{fig:mesh}
\vspace{-1.5em}
\end{figure}
\begin{figure*}[t]
\centering
\includegraphics[width=\linewidth]{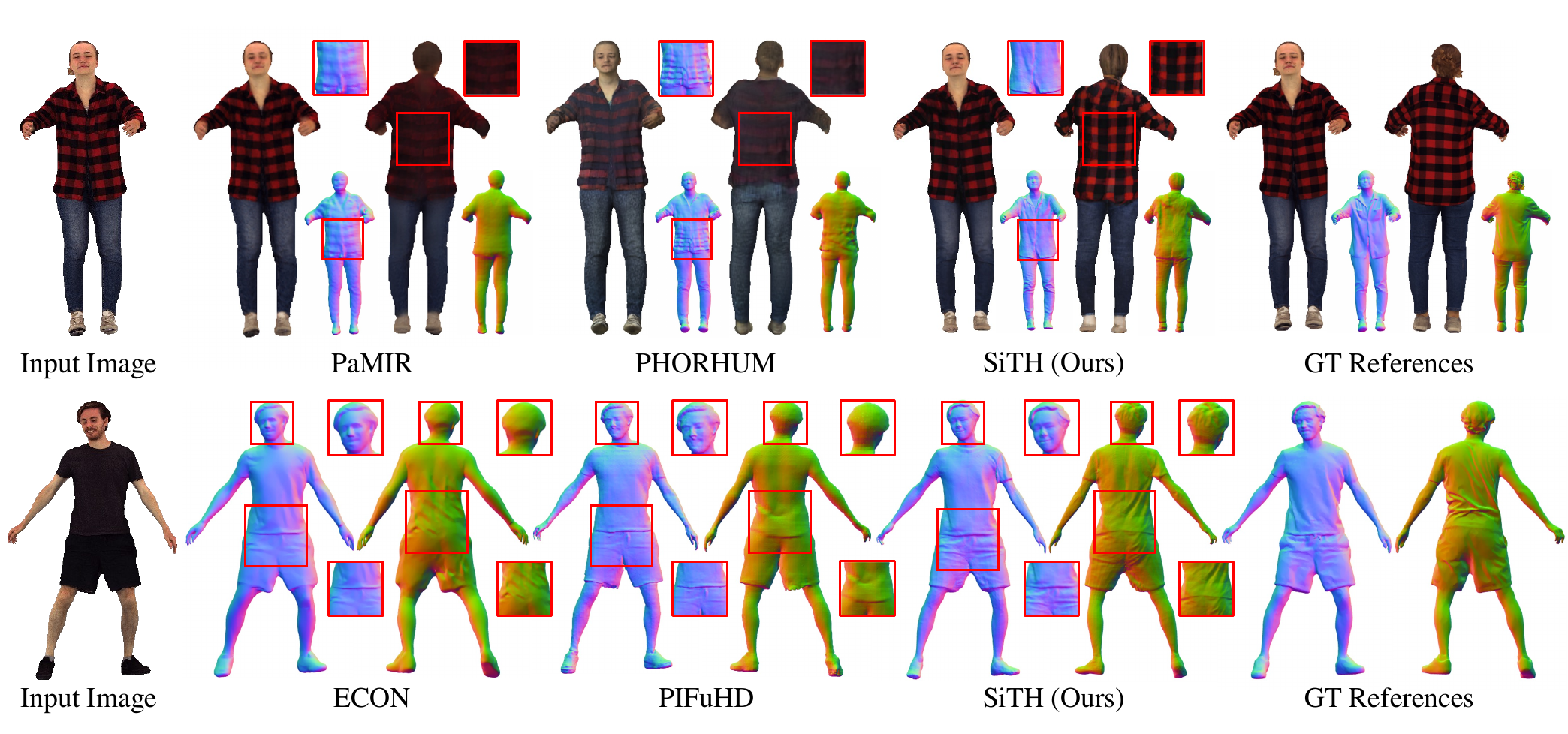}
    \vspace{-2em}
\caption{\textbf{Qualitative comparison on CustomHumans}. \emph{Top}: Results of methods generating mesh and texture. \emph{Bottom}: Results of methods generating mesh only. Note that single-view reconstruction is not possible to replicate exact back-view texture and geometry. Our method generates realistic texture and clothing wrinkles perceptually close to the real scans while other baselines only produce smooth colors and surfaces in the back regions. Best viewed in color and zoom in.}
\label{fig:big}
    \vspace{-1.5em}
\end{figure*}
\subsection{Human Mesh Reconstruction}
\label{sec:mesh}
\vspace{-0.5em}
After obtaining the back-view image, our goal is to construct a full-body human mesh and its textures using the input and back-view image as guidance. We follow the literature~\cite{saito2019pifu,saito2020pifuhd} to model this task with a data-driven method. Given pairwise training data (i.e., front/back-view images and 3D scans), we learn a data-driven model that maps these images to a 3D representation (e.g., a signed distance field (SDF)). We define this mapping as below:
\begin{equation}
\begin{aligned}
      \mathbf{\Phi} : \mathbb{R}^{H\times W\times3} \times \mathbb{R}^{H\times W\times3} \times \mathbb{R}^{3} &\rightarrow \mathbb{R} \times \mathbb{R}^{3}\\
     (I^F, I^B, \mathbf{x}) &\mapsto d_x, \mathbf{r}_x,
\label{eq:mapping}
\end{aligned}
\end{equation}
where $\mathbf{x}$ is the 3D coordinate of a query point, and $d_x, \mathbf{r}_x$ denote the signed distance and RGB color value at point $\mathbf{x}$. The network components we used for learning the mapping function are depicted in~\cref{fig:mesh}.
\vspace{-1.5em}
\paragraph{Local feature querying.}
To learn a generic mapping function that is robust to unseen images, it is important that the model is conditioned solely on local image information with respect to the position of $\mathbf{x}$.
Therefore, we employ the idea of pixel-aligned feature querying~\cite{saito2019pifu,saito2020pifuhd} and separate our model into two branches, i.e., color and geometry. Our model contains a normal predictor that converts the RGB image pair ($I^F, I^B$) into normal maps ($N^F, N^B$). Two image feature encoders $G_d, G_r$ then extract color and geometry feature maps $(\mathbf{f}_d, \mathbf{f}_r) \in \mathbb{R}^{H' \times W' \times D}$ from the images and normal maps respectively (for simplicity we describe the process for a single image and leave out the superscripts, but both front and back images are treated the same).
Finally, we project the query point $\mathbf{x}$ onto the image coordinate (\cref{fig:mesh} red points) to retrieve the local features $(\mathbf{f}_{d,x}, \mathbf{f}_{r,x}) \in \mathbb{R}^D$:
\begin{equation}
\begin{aligned}
      \mathbf{f}_{d,x} &= \mathcal{B}(\mathbf{f}_{d},\pi(\mathbf{x}))= \mathcal{B}(G_d(N),\pi(\mathbf{x})), \\
      \mathbf{f}_{r,x} &= \mathcal{B}(\mathbf{f}_{r},\pi(\mathbf{x}))= \mathcal{B}(G_r(I),\pi(\mathbf{x})),
\end{aligned}
\end{equation}
where $\mathcal{B}$ is a local feature querying operation using bilinear interpolation and $\mathbf{\pi}(\cdot)$ denotes orthographic projection.
\vspace{-1em}
\paragraph{Local positional embedding with skinned body prior. }
As mentioned in~\cref{sec:intro}, a major difficulty in mesh reconstruction is 3D ambiguity where a model has to infer unknown depth information between the front and back images.  
To address this issue, we follow prior work~\cite{zheng2021pamir,xiu2022icon,ho2023custom} leveraging a skinned body mesh~\cite{pavlakos2019expressive} for guiding the reconstruction task. This body mesh is regarded as an anchor that provides an approximate 3D shape of the human body.

To exploit this body prior, we devise a local positional embedding function that transforms the query point $\mathbf{x}$ into the local body mesh coordinate system. We look for the closest point $\mathbf{x}_c^*$ on the body mesh (\cref{fig:mesh} blue point), i.e.,
\begin{equation}
\mathbf{x}_c^* = \arg\min_{\mathbf{x}_c}  \Vert \mathbf{x} - \mathbf{x}_c \Vert_{2},
\end{equation}
where $\mathbf{x}_c$ are points on the skinned body mesh $\mathcal{M}$.
Our positional embedding $\mathbf{p}$ constitutes four elements: a signed distance value $d_c$ between $\mathbf{x}_c^*$ and $\mathbf{x}$, a vector $\mathbf{n}_c= (\mathbf{x} - \mathbf{x}_c^*)$, the UV coordinates $\mathbf{u}_c\in[0,1]^2$ of the point $\mathbf{x}_c^*$, and a visibility label $v_c\in\{1,-1,0\}$ that indicates whether $\mathbf{x}_c^*$ is visible in the front/back image or neither.
Finally, two separate MLPs $H_d, H_r$ take the positional embedding $\mathbf{p}=[d_c,\mathbf{n}_c,\mathbf{u}_c,v_c]$ and the local texture/geometry features $(\mathbf{f}_{d,x}, \mathbf{f}_{r,x})$ as inputs to predict the final SDF and RGB values at point $\mathbf{x}$:
\begin{equation}
\begin{aligned}
    d_x &= H_d (\mathbf{f}_{d,x}^F, \mathbf{f}_{d,x}^B, \mathbf{p}), \\
    \mathbf{r}_x &= H_r(\mathbf{f}_{r,x}^F, \mathbf{f}_{r,x}^B, \mathbf{p}).
\end{aligned}
\end{equation}

\vspace{-1.em}
\definecolor{Gray}{gray}{0.85}
\begin{table*}[t]
    \centering
    \small
\begin{tabular}{l|ccc|c|ccc|c}
\toprule
                     & \multicolumn{4}{c|}{CAPE~\cite{CAPE:CVPR:20}}         & \multicolumn{4}{c}{CustomHuman~\cite{ho2023custom}}          \\
\midrule
Method               & \begin{tabular}[c]{@{}c@{}}CD: P-to-S /\\ S-to-P (cm)$\downarrow$ \end{tabular} & NC$\uparrow$ & f-Score$\uparrow$ & \begin{tabular}[c]{@{}c@{}}LPIPS: F \\ $(\times 10^{-2})\downarrow$ \end{tabular} & \begin{tabular}[c]{@{}c@{}}CD: P-to-S /\\ S-to-P (cm)$\downarrow$ \end{tabular} & NC$\uparrow$ & f-Score$\uparrow$ & \begin{tabular}[c]{@{}c@{}}LPIPS: F / B  \\  $(\times 10^{-2})\downarrow$ \end{tabular} \\
\midrule
 \rowcolor{Gray}PIFu~\cite{saito2019pifu}            & 2.368 / 3.763   & 0.778  & 33.842 & 2.720  & 2.209 / 2.582  & 0.805  & 34.881   & 6.073 / 8.496   \\
 \rowcolor{Gray}PIFuHD~\cite{saito2020pifuhd}        & 2.401 / 3.522  & 0.772 & 35.706   & - & 2.107  / \underline{2.228}   & 0.804 & \textbf{39.076}  & -    \\
 \rowcolor{Gray}PaMIR~\cite{zheng2021pamir}          & \underline{2.190} / \underline{2.806}  & \underline{0.804} & \underline{36.725}  & \underline{2.085} & 2.181  / 2.507  & \underline{0.813} & 35.847 & \underline{4.646} / \underline{7.152}    \\
 \rowcolor{Gray}2K2K~\cite{han2023high}               & 2.478 / 3.683 & 0.782 & 28.700  & - & 2.488 / 3.292  & 0.796  & 30.186  & - \\
 \rowcolor{Gray}FOF~\cite{li2022neurips}               & 2.196 / 4.040 & 0.777 & 34.227  & - & \underline{2.079} / 2.644  & 0.808  & 36.013  & - \\
\midrule
 ICON~\cite{xiu2022icon}               & 2.516  / 3.079  & 0.786  & 29.630 & - & 2.256 / 2.795  & 0.791 & 30.437 & -      \\
 ECON~\cite{xiu2023econ}               & 2.475 / 2.970  & 0.788 & 30.488   & - & 2.483  / 2.680   & 0.797  & 30.894   & -     \\
 SiTH (Ours)                                  & \textbf{1.899 } / \textbf{2.261}  & \textbf{0.816} & \textbf{37.763 }  & \textbf{1.977} & \textbf{1.871} / \textbf{2.045}  &  \textbf{0.826} & \underline{37.029}   
  & \textbf{3.929} / \textbf{6.803}  \\
\bottomrule
\end{tabular}
\caption{\textbf{Single-view human reconstruction benchmarks}. We report Chamfer distance (CD), normal consistency (NC), and f-score between ground truth and predicted meshes. To evaluate texture reconstruction quality, we compute LPIPS between the image rendering of GT and generated textures. \textbf{The best} and \underline{the second best} methods are highlighted in bold and underlined respectively. Note that \colorbox{Gray}{gray color} denotes models trained on more commercial 3D human scans while the others are trained on the public THuman2.0 dataset. }
\label{tab:3D}
\vspace{-1.5em}
\end{table*}

\paragraph{Training and inference.}
We used the same 3D dataset described in~\cref{sec:diffusion} to render training image pairs $(I^F, I^B)$ from the 3D textured scans. For each training scan, query points $\mathbf{x}$ are sampled within a 3-dimensional cube $[-1,1]^3$. For each point, we compute the ground-truth signed distance values $d$ to the scan surface, closest texture RGB values $r$, and surface normal $\mathbf{n}$. Finally, we jointly optimized the normal predictors, the image encoders, and the MLPs in both branches with the following reconstruction losses:
\begin{equation}
    \mathcal{L}_{d} = \Vert d - d_x \Vert_{1} + \lambda_n ( 1 - \mathbf{n} \cdot \nabla_{\mathbf{x}} d_x ),
\end{equation}
\vspace{-1em}
\begin{equation}
    \mathcal{L}_{r} = \Vert \mathbf{r} -  \mathbf{r}_x \Vert_{1}.
\end{equation}
Note that $\nabla_{\mathbf{x}}$ indicates numerical finite differences for computing local normals at point $\mathbf{x}$ and $\lambda_n$ is a hyperparameter.

During inference, we use the input image $I^F$ and the back-view image $\tilde{I}^B$ obtained from~\cref {sec:diffusion} to reconstruct 3D mesh and textures. First, we align both images with the estimated body mesh $\mathcal{M}$ to ensure that image features can be properly queried around the 3D anchor. We adopt a similar strategy of SMPLify~\cite{Bogo:ECCV:2016} to optimize the scale and the offset of the body mesh with silhouette and 2D joint errors. Finally, we perform the marching cube algorithm~\cite{lorensen1998marching} by querying SDF and RGB values within a dense voxel grid via~\cref{eq:mapping} (see~\cref{fig:framework} right).

\begin{figure*}[t]
\centering
\includegraphics[width=\linewidth]{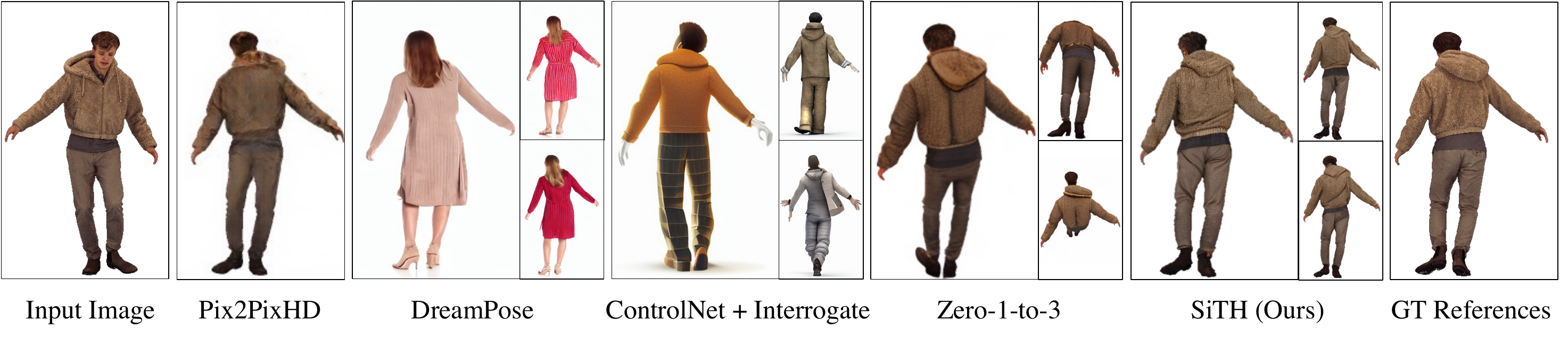}
\vspace{-2em}
\caption{\textbf{Qualitative comparison of back-view hallucination}. We visualize back-view images generated by the baseline methods. Note that the three different images are sampled from different random seeds. Our results are perceptually close to the ground-truth image in terms of appearances and poses. Moreover, our method also preserves generative stochasticity for handling tiny wrinkle changes.}
\label{fig:vs_2D}
\vspace{-1.5em}
\end{figure*}
\vspace{-0.5em}
\section{Experiments}
\label{sec:exp}
\vspace{-0.2em}
\subsection{Experimental Setup}
\vspace{-0.5em}
\label{sec:setup}
\paragraph{Dataset.}
Previous work relied on training data from commercial datasets such as RenderPeople~\cite{renderpeople}.
While these datasets offer high-quality textured meshes, they also limit reproducibility due to limited accessibility.
For fair comparisons, we follow ICON~\cite{xiu2022icon} by training our method on the public 3D dataset THuman2.0~\cite{tao2021function4d} and using the CAPE~\cite{CAPE:CVPR:20} dataset for evaluation.
However, we observed potential biases in the evaluation due to the low-res ground-truth meshes and image rendering defects in the CAPE dataset (for a detailed discussion, please refer to Supp-\cref{sec:supp_dataset}).
Consequently, we further create a new benchmark that evaluates the baselines on a higher-quality 3D human dataset CustomHumans~\cite{ho2023custom}.
In the following, we provide a summary of the datasets used in our experiments:
\begin{itemize}
    \item \textbf{THuman2.0}~\cite{tao2021function4d} contains approximately 500 scans of humans wearing 150 different garments in various poses. We use these 3D scans as the training data.
    \item \textbf{CAPE}~\cite{CAPE:CVPR:20} contains 15 subjects in 8 types of tight outfits. The test set, provided by ICON, consists of 100 meshes. We use CAPE for the quantitative evaluation (\cref{sec:recon}).
    \item \textbf{CustomHumans}~\cite{ho2023custom} contains 600 higher-quality scans of 80 subjects in 120 different garments and varied poses. We selected 60 subjects for all quantitative experiments, user studies, and ablation studies. (\cref{sec:recon} -~\cref{sec:ablation})
\end{itemize}
\vspace{-1.5em}
\paragraph{Evaluation protocol.} We follow the evaluation protocol in OccNet~\cite{mescheder2019occupancy} and ICON~\cite{xiu2022icon} to compute 3D metrics Chamfer distance \textbf{(CD)}, normal consistency \textbf{(NC)}, and \textbf{f-Score}~\cite{tatarchenko2019single} on the generated meshes. To evaluate reconstructed mesh texture, we report \textbf{LPIPS}~\cite{zhang2018unreasonable} of front and back texture rendering.
In user studies, 30 participants rank the meshes obtained by four different methods. We report the \textbf{average ranking} ranging from 1 (best) to 4 (worst).
\subsection{Single-view Human Reconstruction}
\vspace{-.5em}
\paragraph{Benchmark evaluation.} We compared~\methodname~with state-of-the-art single-view human reconstruction methods, including \textbf{PIFu}~\cite{saito2019pifu}, \textbf{PIFuHD}~\cite{saito2020pifuhd},
\textbf{PaMIR}~\cite{zheng2021pamir}, \textbf{FOF}~\cite{li2022neurips}, \textbf{ICON}~\cite{xiu2022icon},  \textbf{PHORHUM}~\cite{alldieck2022phorhum},   \textbf{2K2K}~\cite{han2023high}, and \textbf{ECON}~\cite{xiu2023econ} on CAPE and CustomHumans.
Note that PHORHUM is only used for qualitative comparison since a different camera system is used, leading to the misalignment with ground-truth meshes. 
We visualize the generated mesh texture and normals in~\cref{fig:big}. Existing methods produce over-smoothed texture and normals, particularly in the back.
Our method not only generates photorealistic and perceptually consistent appearances in unobserved regions but also recovers underlying geometric details like clothing wrinkles.

The quantitative results are summarized in~\cref{tab:3D}. 
It's worth noting that most methods are trained with commercial datasets (\colorbox{Gray}{gray color} in~\cref{tab:3D}), while the others are trained on the public THuman2.0 dataset. To evaluate the methods leveraging a skinned body prior (i.e., PaMIR, ICON, ECON, FOF, and SiTH), we use the same pose alignment procedure in their original implementations for a fair comparison.
Results in~\cref{tab:3D} show that the method using a body prior (PaMIR) outperformed the end-to-end method (PIFuHD) on tight clothing and challenging poses in CAPE. However, it falls short in handling diverse outfits in CustomHumans. Moreover, the methods trained on commercial datasets achieve better performance than those trained with public data (ICON, ECON). Notably, our method is robust across both benchmarks, achieving performance comparable to the methods trained on high-quality commercial data.
\begin{figure}[t]
\centering
\includegraphics[width=0.95\linewidth]{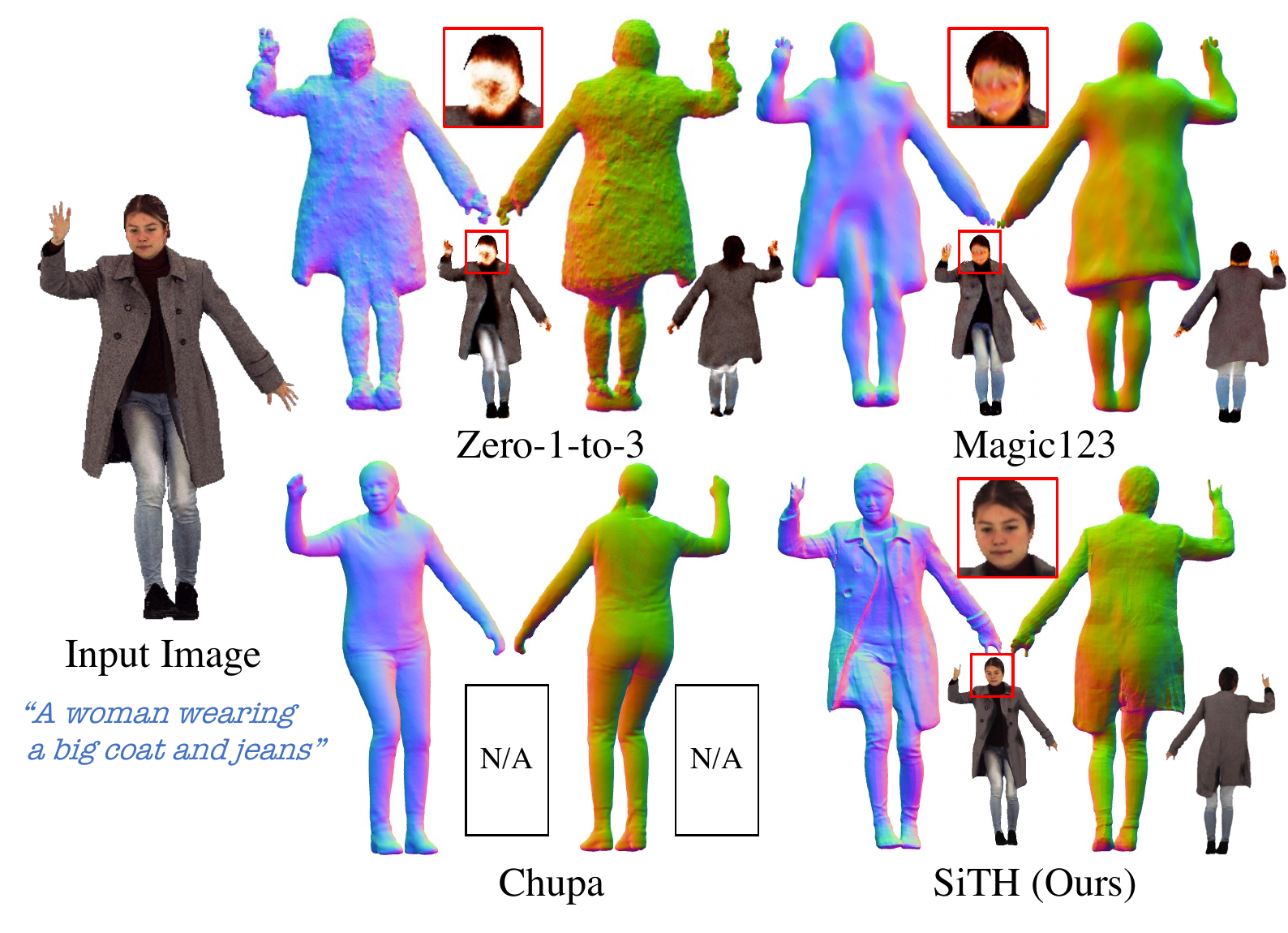}
\vspace{-1em}
\caption{\textbf{Comparison with optimization-based methods}. Compared to methods that utilize diffusion models for optimization, our result is closer to the input image and contains local geometric details. Note that Chupa is not conditioned on the image, and fails to generate correct clothing from the text prompts.}
\label{fig:vs_gen}
\vspace{-1.5em}
\end{figure}
\vspace{-2em}
\paragraph{Compared with optimization-based methods.}
We compared~\methodname~with methods that use pretrained diffusion models and a score distillation sampling loss~\cite{poole2023dreamfusion} to optimize 3D meshes. In the case of \textbf{Zero-1-to-3}~\cite{liu2023zero}, we used the input image to optimize an instant NGP~\cite{mueller2022instant} radiance field, and for \textbf{Magic-123}~\cite{qian2023magic123}, we provided additional text prompts to optimize an SDF tetrahedron~\cite{shen2021dmtet}. From~\cref{fig:vs_gen}, we see that while both methods can handle full-body textures, they struggle with reasoning the underlying geometry and clothing details. It is worth noting that Zero-1-to-3 and Magic-123 require 10 minutes and 6 hours in optimization, respectively, while our method takes under 2 minutes to generate a textured mesh with a marching cube of $512^{3}$ resolution.
Additionally, more similar to our method is \textbf{Chupa}~\cite{kim2023chupa}, which generates front/back-view normals for mesh reconstruction. Note that Chupa is not conditioned on images and does not generate texture. Instead, we provided body poses and text prompts generated by an image-to-text interrogator~\cite{li2022blip} as their conditional inputs. From~\cref{fig:vs_gen}, it's clear that text-conditioning is less accurate than image-conditioning, and the method struggles to generate unseen clothing styles such as coats. By contrast, our method can reconstruct correct clothing geometry and texture from unseen images. We present more discussions and comparisons with optimization-based methods in Supp-\cref{sec:discuss_sds}.
\begin{table}[t]
    \centering
    \small
\begin{tabular}{lcccc}
\toprule
          & ICON & ECON & PIFuHD & Ours   \\
\midrule
 Front Geometry & 3.127 & 2.720 & 2.607 & \textbf{1.547} \\
 Back Geometry  & 3.193 & 2.513 & 3.093 & \textbf{1.200} \\
 Similarity     & 3.093 & 2.660 & 2.780 & \textbf{1.456} \\
\bottomrule
\toprule
          & PIFu & PaMIR & PHOHRUM & Ours   \\
\midrule
 Front Texture  & 3.067 & 2.153 & 3.450 & \textbf{1.327} \\
 Back Texture   & 3.450 & 2.355 & 3.140 & \textbf{1.054} \\
 Similarity     & 3.307 & 2.192 & 3.416 & \textbf{1.093} \\
\bottomrule
\toprule
  &\multicolumn{2}{c}{Chupa} & \multicolumn{2}{c}{Ours}  \\
\midrule
 User Preference &\multicolumn{2}{c}{36.0\%} & \multicolumn{2}{c}{\textbf{64.0\%}}  \\
\bottomrule
\end{tabular}
    \vspace{-.5em}
\caption{\textbf{User study results}. \emph{Top}: 30 users are asked to rank the quality of surface normal images from best (1) to worst (4). We report the average ranking of each method. \emph{Middle}: Similar to the first task, users are asked to rank the quality of RGB textures. \emph{Bottom}: We ask users to choose the mesh with a better quality.}
    \vspace{-2em}
\label{tab:user}
\end{table}

\vspace{-1.5em}
\paragraph{User study.} The above metrics may not fully capture the quality of 3D meshes in terms of realism and local details.
To address this, we conducted a user study to compare the texture and geometry quality among various baselines.
We invited 30 users to rank the front/back-view texture and normal renderings of 3D meshes generated by four different methods. Additionally, we asked the users to assess the similarity between the input images and the generated meshes. The results (\cref{tab:user}) support our claim that existing methods struggle to efficiently generate desirable back-view texture/geometry from single-view images.
Our method, which leverages the generative capability of diffusion models, consistently outperforms each baseline. 
It also produces more preferred front-view textures and geometries, as evidenced by higher user rankings.
We also conducted a user study with Chupa (in~\cref{tab:user} \emph{bottom}) which also indicates more users prefer the 3D meshes generated by our method.

\label{sec:recon}
\vspace{-.5em}
\subsection{Generative Capability}
\label{sec:generative}
\vspace{-0.5em}
\paragraph{Image quality comparison.}
Our hallucination module is a unique and essential component that generates spatially aligned human images to guide 3D mesh reconstruction.
Given that our focus is on back-view hallucination, we compare the quality of generated images with the relevant generative methods in~\cref{fig:vs_2D}.
We trained a baseline \textbf{Pix2PixHD}~\cite{wang2018pix2pixHD} model, which produced smooth and blurry results on unseen images due to overfitting to 500 subjects.
Another method closely related to ours is \textbf{DreamPose}~\cite{dreampose_2023}, which conditions the model with DensePose images and finetunes the diffusion model with paired data. However, their model failed to handle unseen images, in contrast to our approach.
While \textbf{Zero-1-to-3}~\cite{liu2023zero} can generalize to unseen images, their method faces challenges in generating consistent body poses given the same back-view camera.
Moreover, we designed another baseline that provides \textbf{ControlNet}~\cite{zhang2023adding} for corresponding text prompts using an image-to-text interrogator~\cite{li2022blip}.
However, without proper image conditioning and fine-tuning, such a method cannot generate images that faithfully match the input appearances.
Our method not only addresses these issues but also handles stochastic appearances (e.g., tiny differences in wrinkles) from different random seeds. We report 2D generative evaluation metrics and more results in Supp-\cref{sec:compare2D_supp}.

\begin{table}[t]
    \centering
    \small
\begin{tabular}{lccc}
\toprule
 Method               & CD (cm)$\downarrow$ & NC$\uparrow$ & f-Score$\uparrow$     \\
\midrule
 W/o Body Mesh     & 2.471  & 0.801 & 33.244   \\
 W/o Hallucination  & 1.960  & \textbf{0.840} & 36.677    \\
 Full Pipeline      & \textbf{1.958}  & 0.826 & \textbf{37.029}    \\
 \midrule
 W/ GT Body Mesh    & 1.172 & 0.891 & 58.858    \\
 W/ GT Body and $I^B$   & 1.059 & 0.914 & 63.356      \\
\bottomrule
\end{tabular}
\caption{\textbf{Ablation study on CustomHumans}. We ablate the hallucination module and the skinned body mesh in our pipeline. Please refer to our discussion in~\cref{sec:ablation}.}
\vspace{-2em}

\label{tab:ablation}
\end{table}

\vspace{-.5em}
\subsection{Ablation Study}
\begin{figure}[t]
\centering
\includegraphics[width=\linewidth]{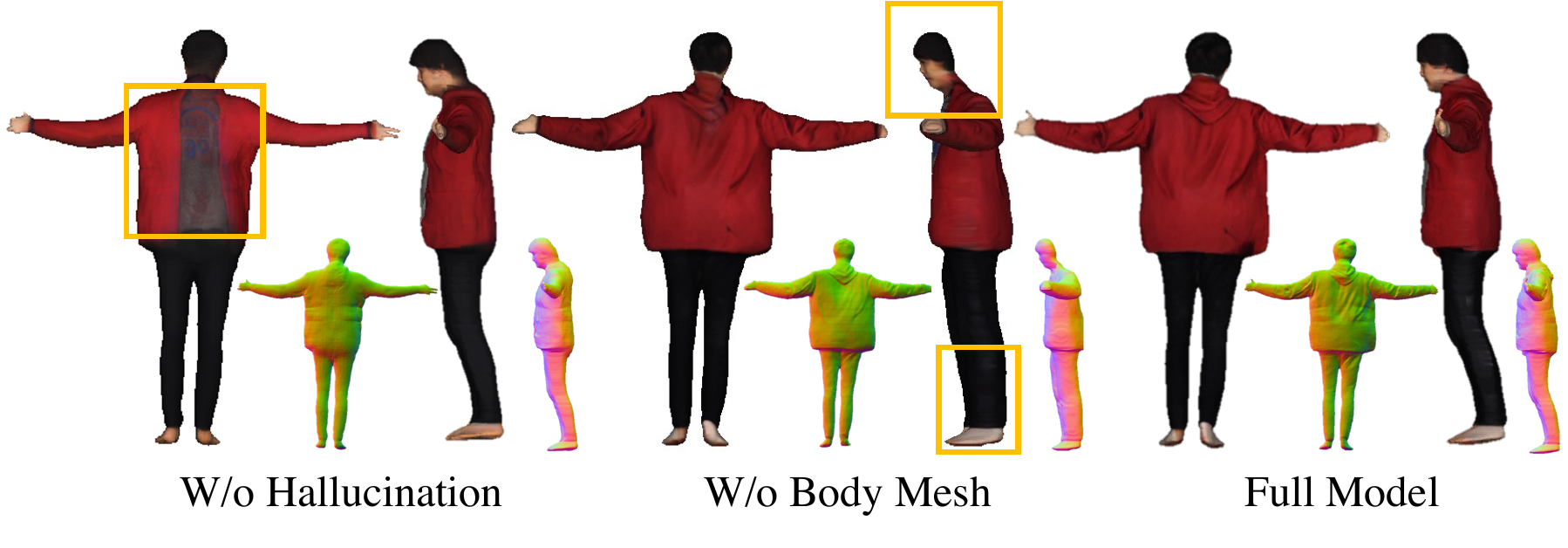}
\vspace{-2em}
\caption{\textbf{Ablation study}. We visualize back and side-view rendering of the reconstructed meshes. Our full model produced a correct body shape and more realistic clothing geometry. }
\label{fig:ablation}
\vspace{-1.em}
\end{figure}

\vspace{-0.5em}
\label{sec:ablation}
We conducted controlled experiments to validate the effectiveness of our proposed modules. As shown in~\cref{fig:ablation}, the skinned body mesh is a crucial component for 3D human reconstruction. Without this body mesh as a 3D anchor, the output mesh contains an incorrect body shape due to the depth ambiguity issue. Conversely, removing the hallucination module has minimal impact on 3D reconstruction metrics, though it slightly degrades normal consistency. However, the overall quality in both texture and geometry is incomparable with our full model (see~\cref{fig:ablation}~\emph{right}). This is consistent with our findings in user studies, indicating that 3D metrics may not accurately reflect the perceptual quality of 3D meshes.
Finally, we tested two additional variants, leveraging ground-truth body meshes and real back-view images in our full pipeline, representing the upper bound of our method. As shown in~\cref{tab:ablation} \emph{bottom}, this additional information notably improves the 3D metrics. These results highlight the persistent challenges in the single-view reconstruction problem, including pose ambiguity and the stochastic nature of clothing geometry. For more experiments on our design choices, please refer to Supp-\cref{sec:design_supp}.
\vspace{-1.5em}
\subsection{Applications}
\label{sec:app}

\begin{figure}[t]
\centering
\includegraphics[width=\linewidth]{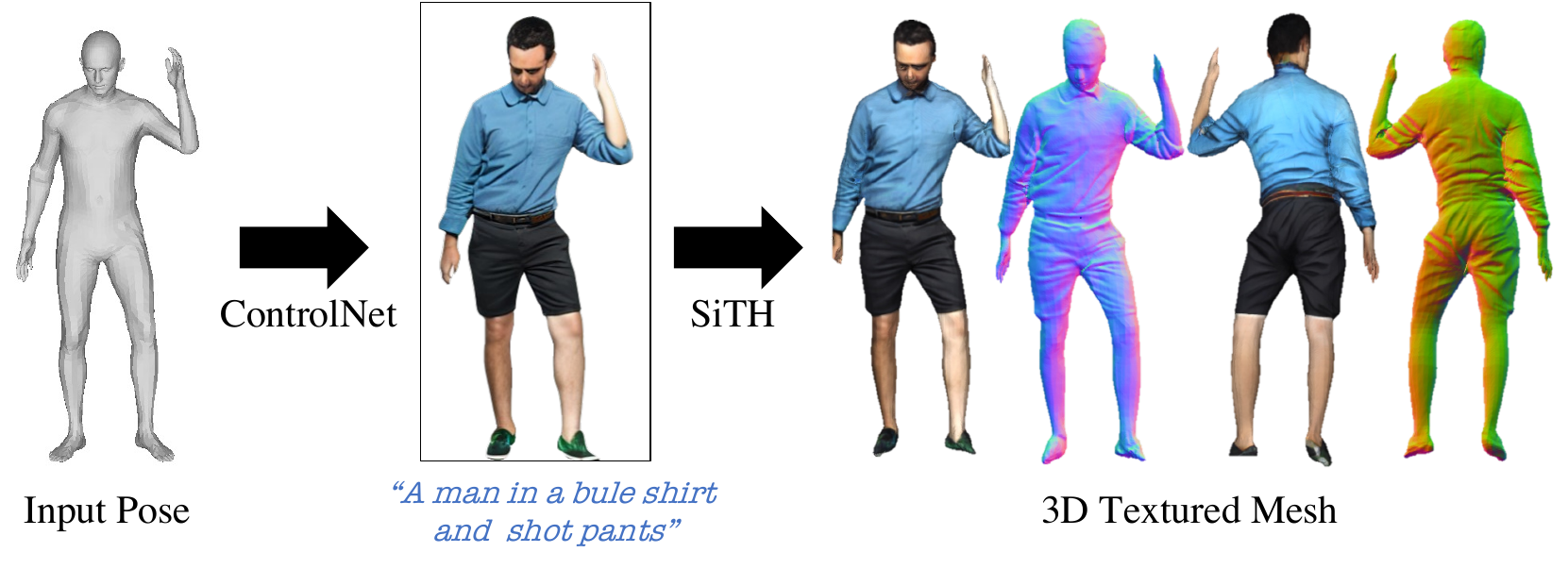}
\vspace{-2em}
\caption{\textbf{Applications}. Our pipeline can be incorporated with text-to-image diffusion models for 3D human creation.
}
\label{fig:app}
\vspace{-2em}
\end{figure}
\vspace{-0.5em}
Inheriting the generative capability of LDM,~\methodname~is robust to diverse inputs, such as out-of-distribution or AI-generated images. We demonstrate a unique solution to link photo-realistic AI photos and high-fidelity 3D humans.
In~\cref{fig:app}, we introduce a 3D creation workflow integrating powerful text-to-image generative models.
Given a body pose, we generate a front-view image using Stable Diffusion and ControlNet using text prompts. \methodname~then creates a full-body textured human from the AI-generated image. 
%handle out-of-distribution and unseen input images. For instance, we hallucinate the back-view images of a game character and create a 3D model from the image.
%To highlight this characteristic, we showcase two applications leveraging our proposed single-view reconstruction pipeline. 

%\paragraph{Zero-shot adaptation.} In~\cref{fig:app} (\emph{top}), we show that our pipeline can handle out-of-distribution and unseen input images. For instance, we hallucinate the back-view images of a game character and create a 3D model from the image. Our method does not require domain-specific data for retraining the model, demonstrating its adaptability in various domains. 

%\paragraph{Mesh editing.} In~\cref{fig:app} (\emph{bottom}), we demonstrate an integration of our method with the existing powerful text-to-image diffusion models. Given a 3D mesh and its underlying SMPL-X body mesh, we generate a new front-view image using Stable Diffusion and ControlNet models with text prompts given by users. Our method creates a new full-body 3D model that has the same body pose as the 3D mesh but with a new appearance from the AI-generated images. Our method uniquely bridges the gap between photo-realistic AI photos and high-fidelity 3D human models. 
\vspace{-1.em}
\section{Conclusion}
\vspace{-0.5em}
\label{sec:con}
We propose an innovative pipeline designed to create fully textured 3D humans from single-view images.
Our approach seamlessly integrates an image-conditioned diffusion model into the existing data-driven 3D reconstruction workflow.
Leveraging the generative capabilities of the diffusion model, our method efficiently produces lifelike 3D humans from a diverse range of unseen images in under two minutes. We expect our work will advance the application of generative AI in 3D human creation.

\paragraph{Acknowledgements.} This work was partially supported by the Swiss SERI Consolidation Grant "AI-PERCEIVE". We thank Xu Chen for insightful discussions, Manuel Kaufmann for suggestions on writing
and the title, and Christoph Gebhardt, Marcel Buehler, and
Juan-Ting Lin for their writing advice.
\addtocontents{toc}{\protect\setcounter{tocdepth}{2}}

{
    \small
    \bibliographystyle{ieeenat_fullname}
    \bibliography{main}
}

\clearpage

\maketitlesupplementary

{
  \hypersetup{linkcolor=black}
  \tableofcontents 
}

\noindent\rule{\linewidth}{0.4pt}

\section{Benchmark Description}
\label{sec:supp_dataset}
\begin{figure*}[t]
\centering
\includegraphics[width=0.8\linewidth]{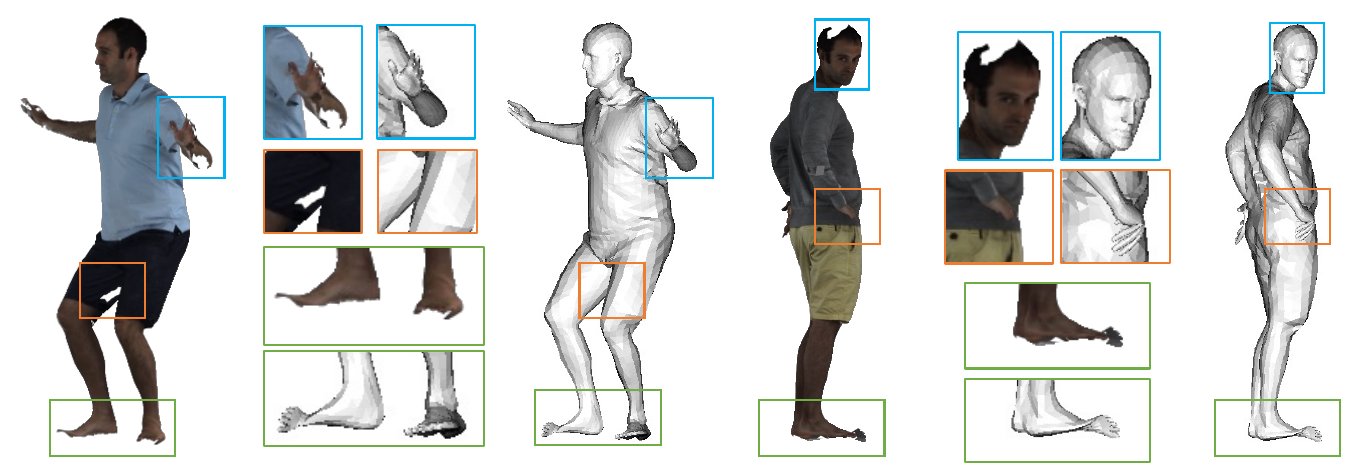}
\caption{\textbf{Defects in the CAPE dataset}. The rendering defects from incomplete point clouds result in a notable discrepancy between the input images and the ground-truth meshes in the CAPE dataset.}
\label{fig:cape_supp}
\vspace{-.5em}
\end{figure*}

We provide detailed descriptions of the CAPE~\cite{CAPE:CVPR:20} and the CustomHumans~\cite{ho2023custom} dataset used for benchmark evaluation in our study. The CAPE dataset includes sequences of posed humans featuring 15 subjects. For evaluation purposes, ICON~\cite{xiu2022icon} selected 100 frames, each consisting of RGBA images from three viewpoints and an SMPL+D (vertex displacements) ground-truth mesh. We identified several limitations in the CAPE dataset (refer to~\cref{fig:cape_supp}) Firstly, there is limited diversity in human outfits, as most subjects wear tight clothing such as t-shirts and shorts. Secondly, the images are rendered from unprocessed point clouds, leading to rendering defects. Lastly, the ground-truth meshes are of low resolution and do not fully correspond to the input images. These issues suggest that experiments conducted solely on the CAPE dataset may be biased.

\begin{figure}[t]
\centering
\includegraphics[width=\linewidth]{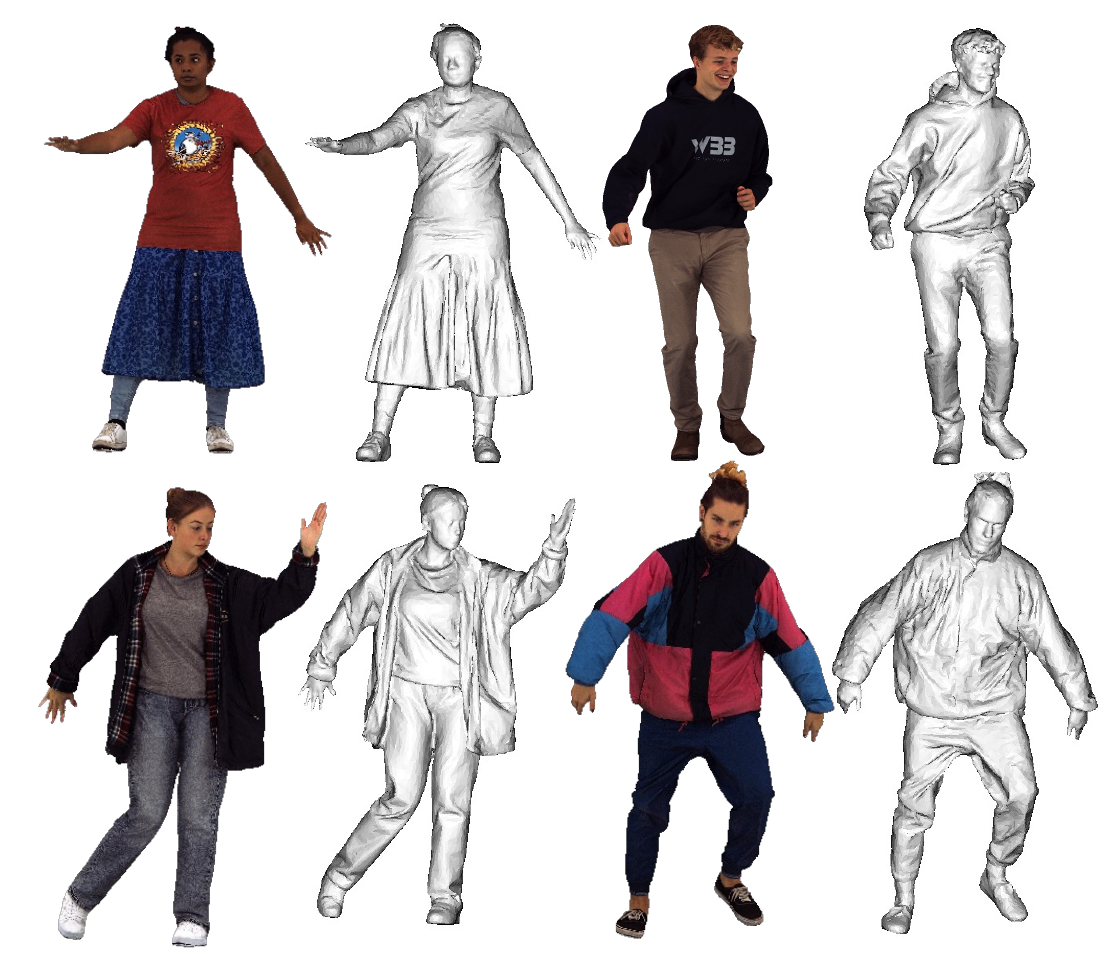}
\caption{\textbf{Examples of images and ground-truth scans in CustomHumans}. Our new benchmark contains diverse and challenging human scans for evaluation.}
\label{fig:custom_supp}
\vspace{-1.5em}
\end{figure}

To ensure an unbiased evaluation, we introduced a new benchmark using the higher-quality, publicly available 3D human dataset, CustomHumans~\cite{ho2023custom}. Specifically, we selected 60 textured human scans, each featuring different outfits, for evaluation. For each scan, we rendered test images from four different viewpoints. Note that we directly rasterize the textured scans to obtain the input images, ensuring that the ground-truth mesh precisely corresponds to the images.~\cref{fig:custom_supp} showcases samples from our benchmarks, highlighting the increased diversity of the clothing.


\section{Implementation Detail}
\subsection{Back-view Hallucination Module}
We detail the implementation of our image-conditioned diffusion model described in~\cref{sec:diffusion}. Our model backbone is based on the Stable Diffusion image variations~\cite{sd_image} which leverages CLIP features for cross-attention and VAE features for concatenation in image conditioning. In both training and inference, the pretrained VAE autoencoder and the CLIP image encoder are kept frozen. We initialize the diffusion U-Net’s weights using the Zero-1-to-3~\cite{liu2023zero} model and create a trainable ControlNet~\cite{zhang2023adding} model following the default network setups but with an adjustment to the input channels. The ControlNet inputs contain 4 channels of masks and UV images with an optional 4 channels of camera view angles. The camera view angles are essential only when generating images from arbitrary viewpoints (instead of only back-view). The ControlNet model and the diffusion U-Net’s cross-attention layers are jointly trained with $512\times 512$ resolution multi-view images, rendered from the THuman2.0 dataset~\cite{tao2021function4d}.

For each scan in THuman2.0, we render front-back image pairs from 20 camera angles, resulting in around 10k training pairs. We also randomly change the background colors for data augmentation. For training, we utilize a batch size of 16 images and set the learning rate to $4 \times 10^{-6}$ incorporating a constant warmup scheduling. The ControlNet model’s conditioning scale is fixed at 
$1.0$. We employ classifier-free guidance in our training, which involves a dropout rate of $0.05$ for the image-conditioning. The training takes about two days on one NVIDIA A100 GPU for 10k steps. During inference, we apply a classifier-free guidance scale of $2.5$ to obtain the final output images.  

\subsection{Mesh Reconstruction Module}
We follow the methodology of PIFuHD~\cite{saito2020pifuhd}, using the HourGlass~\cite{newell2016stacked} and the fully convolutional~\cite{johnson2016perceptual} model as our image feature extractors and the normal predictor, respectively. 
The feature extractors yield a 32-dimensional feature map for feature querying.  
Our geometry MLP is designed with five layers of 512-dimensional linear layers, each followed by a leakyReLU activation function. Skip connections are applied at the third, fourth, and fifth layers.
On the other hand, the texture MLP comprises four layers of 256-dimensional linear layers, with skip connections at the third and fourth layers.

We first train the normal predictor using normal images rendered from the THuman2.0 dataset.
We optimize the normal predictor with an L1 reconstruction loss for 600 epochs.
Subsequently, we proceed to jointly train the feature extractor and the SDF MLPs with a learning rate of $0.001$ and a batch size of 2 scans.
The normal predictor is jointly fine-tuned with a learning rate $1 \times 10^{-5}$.
We set the hyperparameter $\lambda_n$ to $0.1$.
During each training iteration, we sample 40,960 query points within a thin shell surrounding the ground-truth mesh surfaces. The entire training process requires approximately five days on a single NVIDIA A100 GPU for 800 epochs on the THuman2.0 dataset. Finally, we train the other feature extractor and the RGB MLPs with a learning rate of $0.001$ and a batch size of 2 scans for 200 epochs. During inference, a 3D textured mesh can be reconstructed under two minutes with an NVIDIA 3090 GPU. This includes pose estimation and mask prediction (3s), generation of a back-view image (4.5s), alignment of the body mesh and the input images (10s), and mesh reconstruction at the marching cube resolution of $512^3$ (60s).

\section{More Experimental Results}
\label{sec:supp_exp}

\subsection{Image Quality Comparison}
\label{sec:compare2D_supp}
\begin{figure*}[t]
\centering
\includegraphics[width=\linewidth]{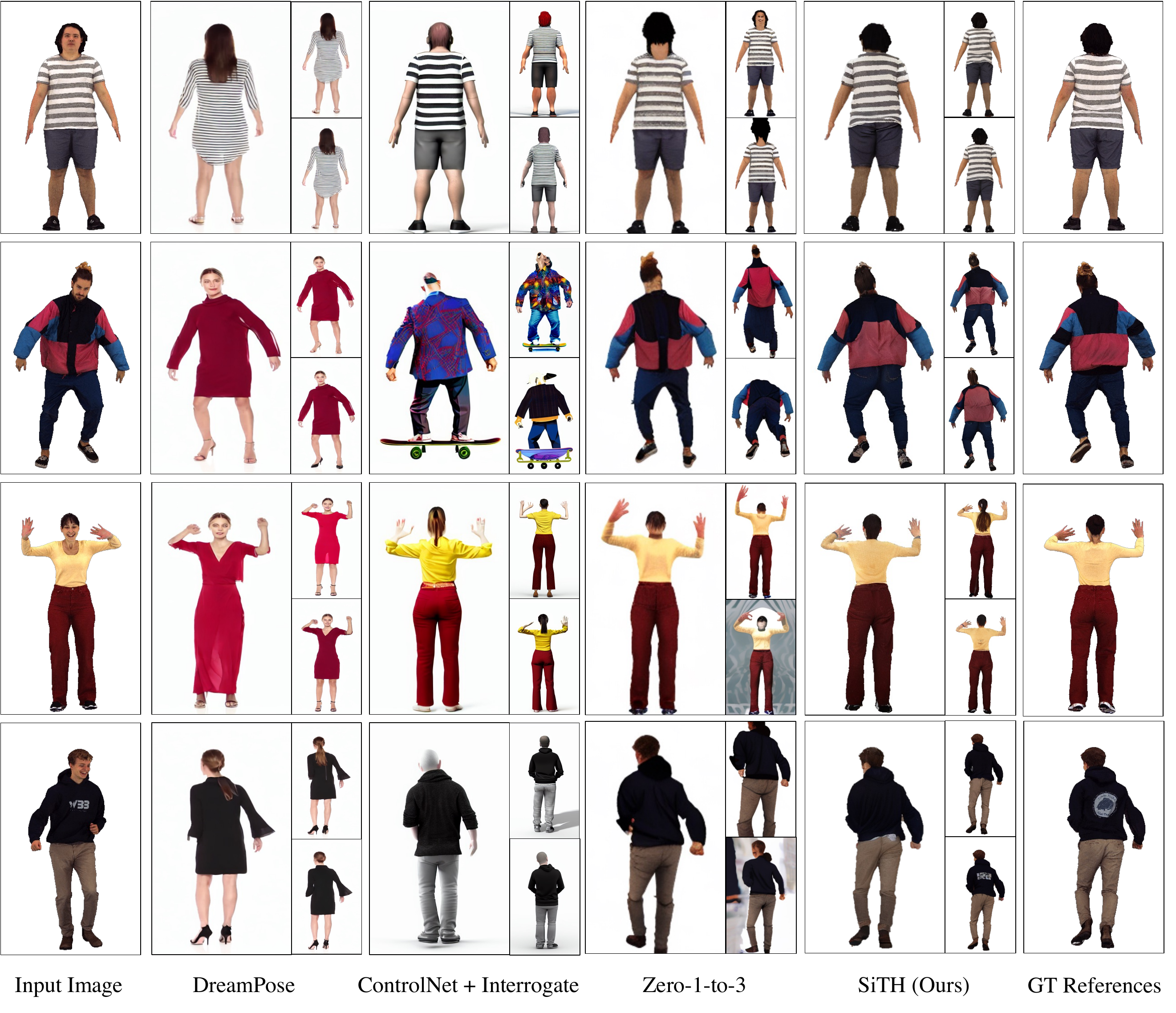}
\vspace{-2em}
\caption{\textbf{Qualitative comparison of back-view hallucination}. We visualize back-view images generated by the baseline methods. Note that the three different images are sampled from different random seeds. Our results are perceptually close to the ground-truth image in terms of appearances and poses. Moreover, our method also preserves generative stochasticity for handling hairstyles and clothing colors.}
\label{fig:vs_2D_supp}
\vspace{-.5em}
\end{figure*}
\begin{table}[t]
    \centering
    \small
\resizebox{\columnwidth}{!}{%
\begin{tabular}{lcccc}
\toprule
 Method               & SSIM$\uparrow$ & LPIPS$\downarrow$ & \begin{tabular}[c]{@{}c@{}}KID\\ ($\times 10^{-3}$)$\downarrow$ \end{tabular}& \begin{tabular}[c]{@{}c@{}}Joints Err.\\ (pixel)$\downarrow$ \end{tabular}  \\
\midrule
 Pix2PixHD~\cite{wang2018pix2pixHD} & 0.816 & 0.141 & 86.2 & 53.1 \\
 DreamPose~\cite{dreampose_2023}        & 0.844   & 0.132 & 86.7 & 76.7   \\
Zero-1-to-3~\cite{liu2023zero}         & 0.862   & 0.119 & 30.0  & 73.4    \\
 \begin{tabular}[l]{@{}l@{}}ControlNet~\cite{zhang2023adding}\\ +Interrogate\end{tabular}  & 0.851   & 0.202 & 39.0 & 35.7    \\
 \methodname~(Ours)             & \textbf{0.950}   & \textbf{0.063} & \textbf{3.2}  & \textbf{21.5}    \\

\bottomrule
\end{tabular}
}
\caption{\textbf{Hallucination comparison on CustomHumans}. We compute image metrics between the generated and ground-truth back-view images. Our method achieved the best image quality and pose accuracy.
}
\vspace{-1.5em}
\label{tab:2D_supp}
\end{table}

We carry out a quantitative evaluation on the images generated by \textbf{Pix2PixHD}~\cite{wang2018pix2pixHD}, \textbf{DreamPose}~\cite{dreampose_2023}, \textbf{Zero-1-to-3}~\cite{liu2023zero}, \textbf{ControlNet}~\cite{zhang2023adding}, using ground-truth back-view images for comparison (see~\cref{tab:2D_supp}). To assess the image quality, we employ various metrics, including multi-scale Structure Similarity (\textbf{SSIM})~\cite{wang2003multiscale}, Learned Perceptual Image Patch Similarity (\textbf{LPIPS})~\cite{zhang2018unreasonable}, Kernel Inception Distance (\textbf{KID})~\cite{kid}, and \textbf{2D joint errors} using a pose predictor~\cite{mediapipe}. Our method demonstrates better performance over the others in terms of similarity, quality, and pose accuracy.

In~\cref{fig:vs_2D_supp}, we present additional results generated by these methods. DreamPose exhibits overfitting issues, failing to accurately generate back-view images with the correct appearances.
Although ControlNet successfully predicts images with correct poses, it shows less accuracy in text conditioning, particularly in generating inconsistent appearances. 
Zero-1-to-3, shows instability in view-point conditioning, resulting in a noticeable variance in the human body poses in the generated images. 
In contrast, our method not only produces more faithful back-view images but also handles stochastic elements such as hairstyles and clothing colors.

\subsection{3D Reconstruction Plugin}
\label{sec:plugin_supp}

\begin{figure}[t]
\centering
\includegraphics[width=\linewidth]{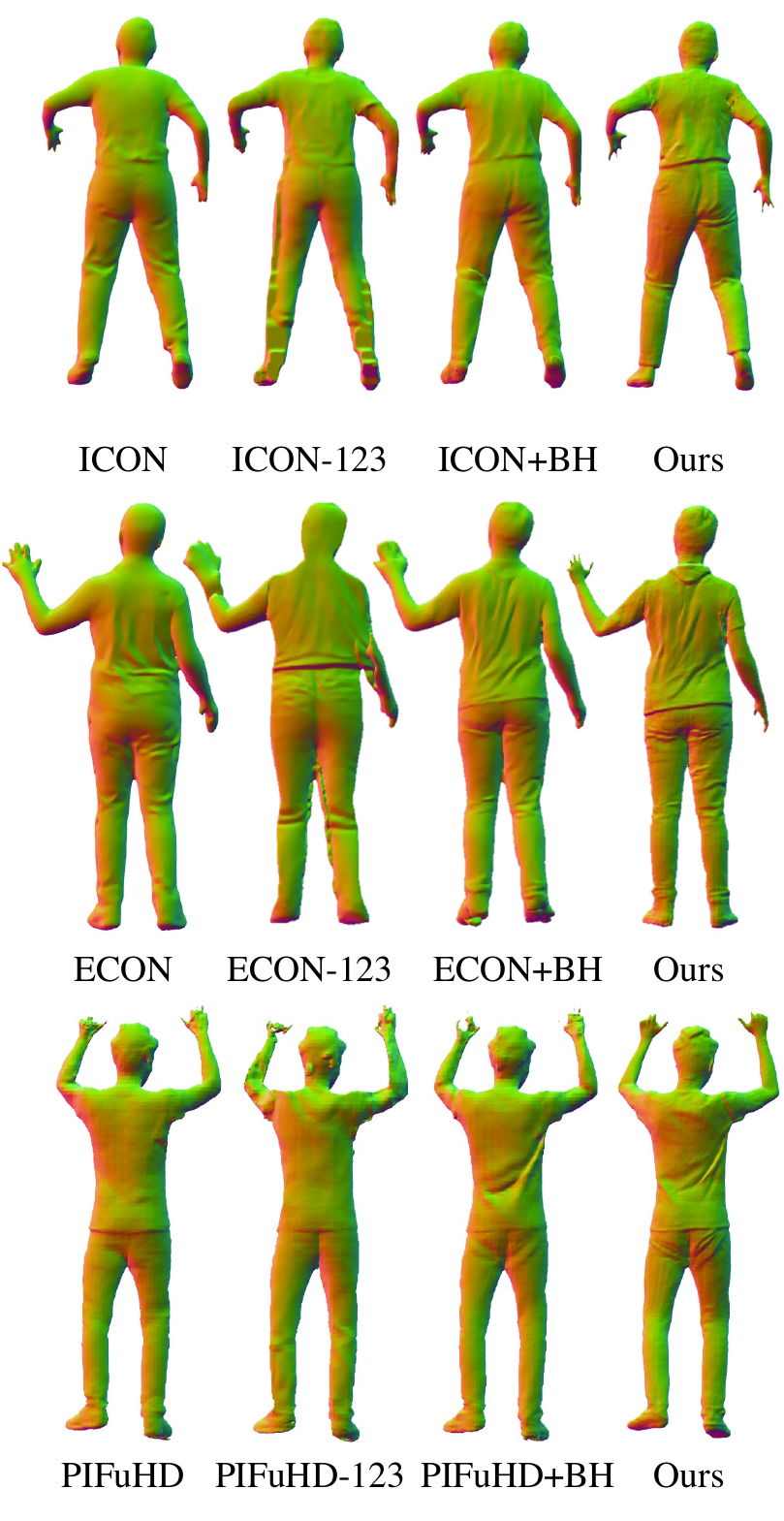}
\caption{\textbf{Reconstruction plugin}. We replaced the back normal images typically used in existing 3D reconstruction methods with our generated back-view images. This modification enhances the perceptual qualities of these baseline methods.}
\label{fig:plugin}
\vspace{-1.5em}
\end{figure}

We demonstrate that our hallucination module can be seamlessly integrated into existing single-view clothed human reconstruction pipelines.
We implemented variants of ICON, ECON, and PIFuHD by providing them back normal from our generated back-view images (denoted as \textbf{+BH}). These are then compared to the original methods and their respective variants using the Zero-1-to-3 model as a plugin (denoted as \textbf{-123}). 
As shown in~\cref{fig:plugin}, integrating Zero-1-to-3 with these methods did not produce satisfactory clothing geometry. In contrast, our hallucination module yielded more realistic clothing wrinkles and enhanced the perceptual quality of ICON, ECON, and PIFuHD.
Note that even though we provide additional images with these baselines, our pipeline still produced more detailed geometry and correct body shapes. This again verifies the importance and effectiveness of our mesh reconstruction module. 

The quantitative results, presented in Table~\cref{tab:plugin}, further support these findings. We observed that the combination of Zero-1-to-3 with these methods did not lead to significant improvements. 
However, our hallucination module slightly enhanced the 3D metrics for ICON and ECON but had a marginally negative impact on PIFuHD. 
The reason can be observed from~\cref{fig:plugin} where PIFuHD tends to produce smooth surfaces that result in better numeric performance.
ICON and ECON benefit from our hallucinations since their original model produced artifacts and incorrect clothing details. 
This finding also confirms the necessity of our user studies in~\cref{sec:recon} since the visual quality is hard to measure by the existing metrics.

\begin{table}[t]
    \centering
    \small
\begin{tabular}{l|ccc}
\toprule
 Method       & \begin{tabular}[c]{@{}c@{}}Pred-to-Scan /\\ Scan-to-Pred (mm)$\downarrow$ \end{tabular} & NC$\uparrow$ & f-Score$\uparrow$     \\
 \midrule
 ICON~\cite{xiu2022icon}         & 22.562 / 27.954  & 0.791 & 30.437    \\
 ICON-123     & \red{-0.716} / \red{-3.478}  & \blue{-0.004} & \red{+2.991} 
 \\
 ICON+BH        & \red{-1.208} / \red{-3.728}  & \red{+0.014} & \red{+3.831} \\
\midrule
 ECON~\cite{xiu2023econ}         & 24.828 / 26.802 & 0.797 & 30.894    \\
 ECON-123     & \blue{+0.646} / \blue{+1.115} & \blue{-0.024} &	\blue{-1.273} \\
 ECON+BH        & \blue{+0.001} / \blue{+0.797} & \blue{-0.003} & \blue{-0.175}  \\ 
 \midrule
 PIFuHD~\cite{saito2020pifuhd}      & 21.065 / 22.278 & 0.804 & 39.076   \\
 PIFuHD-123   & \blue{+1.433} / \blue{+2.477} & \blue{-0.034} &	\blue{-3.139}  \\
 PIFuHD+BH      & \blue{+0.418} / \red{-0.589} & \blue{-0.005} &	\red{+0.066}      \\
\bottomrule
\end{tabular}
\caption{\textbf{Generative plugins for 3D reconstruction}. We extend the baseline methods with Zero-1-to-3 (denoted as -123) and our hallucination module (denoted as +BH). Our method improves their perceptual qualities without affecting their overall performance. \red{Red} and \blue{blue} indicate improvements and decreases respectively. }
\vspace{-1.5em}
\label{tab:plugin}
\end{table}

\definecolor{Gray}{gray}{0.85}

\begin{table*}[t]
    \centering
    \small
\begin{tabular}{l|ccc|ccc}
\toprule
                     & \multicolumn{3}{c|}{CAPE~\cite{CAPE:CVPR:20}}         & \multicolumn{3}{c}{CustomHumans~\cite{ho2023custom}}          \\
\midrule
Method               & \begin{tabular}[c]{@{}c@{}}Pred-to-Scan /\\ Scan-to-Pred (mm)$\downarrow$ \end{tabular} & NC$\uparrow$ & f-Score$\uparrow$  & \begin{tabular}[c]{@{}c@{}}Pred-to-Scan /\\ Scan-to-Pred (mm)$\downarrow$ \end{tabular} & NC$\uparrow$ & f-Score$\uparrow$  \\
\midrule
 \rowcolor{Gray}PIFu~\cite{saito2019pifu}     & 26.359 / 40.642   & 0.755 & 29.283 & 24.765 / 34.007  & 0.780 & 31.911  \\
 \rowcolor{Gray}PIFuHD~\cite{saito2020pifuhd}   & 25.644 / 38.050  & 0.755 & 32.157 & 23.004 / 30.039  & 0.785 & \textbf{36.311}  \\
 \rowcolor{Gray}FOF~\cite{li2022neurips}        & \underline{21.671} / 37.246  & 0.778 & \underline{33.971} & \underline{21.995} / 31.076  & 0.789 & 34.403   \\
 \rowcolor{Gray}PaMIR~\cite{zheng2021pamir}     & 24.737 / \underline{33.049}  & \underline{0.782}  & 31.621  & 23.471  / \underline{30.023}  & \underline{0.797} & 34.404 \\
 \midrule
 ICON~\cite{xiu2022icon}     & 27.897 / 36.907  & 0.757 & 25.898 & 25.957 / 37.857  & 0.763 & 26.857  \\
ECON~\cite{xiu2023econ}     & 27.333  / 34.364   & 0.765 & 26.960 & 27.447 / 38.858   & 0.757 & 27.075   \\
 \methodname~(Ours)    & \textbf{21.324} / \textbf{29.050} & \textbf{0.791} & \textbf{34.199} & \textbf{20.513} / \textbf{28.923}  &  \textbf{0.804} & \underline{35.824} \\
\bottomrule
\end{tabular}
\caption{\textbf{Single-view human reconstruction from multiple viewpoints}. We report Chamfer distance, normal consistency (NC), and f-score between ground truth and predicted meshes. Note that \colorbox{Gray}{gray color} denotes models trained on more commercial 3D human scans while the others are trained on with the public THuman2.0 dataset.}
\vspace{-.5em}

\label{tab:3D_supp}
\end{table*}

\definecolor{Gray}{gray}{0.85}

\begin{table*}[t]
    \centering
    \small
\begin{tabular}{l|ccc|ccc}
\toprule
 Angle       & \begin{tabular}[c]{@{}c@{}}Pred-to-Scan /\\ Scan-to-Pred (mm)$\downarrow$ \end{tabular} & NC$\uparrow$ & f-Score$\uparrow$  & $\Delta$ CD &  $\Delta$ NC & $\Delta$ f-Score  \\
 \midrule
 $0^{\circ}$      & 16.880 / 20.314 & 0.8423  & 39.850 & - & - & - \\ 
 $15^{\circ}$     & 16.428 /	20.177 & 0.8428  & 39.971 & -0.452 / -0.137 & +0.0005 & +0.121\\
 $30^{\circ}$     & 17.806 / 22.802 & 0.8305	& 37.154 & +0.926 / +2.488 & -0.0118  & -2.696\\
 $45^{\circ}$     & 18.585 /	23.308 & 0.8243  & 35.652 & +1.705 / +2.994 & -0.0180 & -4.198\\
 \rowcolor{Gray}$60^{\circ}$     & 20.404 /	29.519 & 0.8052  & 33.675 & +3.524 / \textbf{+9.205} & -0.0371 & -6.175\\
 $75^{\circ}$     & 22.111 /	33.309 & 0.7960  & 32.334 & +5.231 / +12.995 & -0.0463 & -7.516\\
 $90^{\circ}$     & 23.752 / 38.338 & 0.7816  & 30.011 & +6.872 / +18.024 & -0.0607 & -9.839\\

\bottomrule
\end{tabular}
\caption{\textbf{Robustness of 3D reconstruction with respect to view angles}. We tested our pipeline using the images and textured scans that were rotated by varying view angles. Note that we use GT back-view images and only analyze the robustness of the mesh reconstruction module. The results from these tests demonstrate that our method maintains robustness within a view angle change of up to 45 degrees. }

\label{tab:angle_supp}
\vspace{-.5em}
\end{table*}

\subsection{More Benchmark Evaluation}
\label{sec:eval_supp}

We present detailed descriptions of our benchmark evaluation protocol. For a fair comparison, we generated meshes from all baselines using marching cubes with a resolution of 256. 
To accurately compare the reconstructed meshes with ground-truth meshes, we utilize the Iterative Closest Point (ICP) algorithm~\cite{besl1992method} to register reconstructed meshes. 
This step is crucial for aligning the meshes with ground truth, thereby eliminating issues of scale and depth misalignment of different methods.
When calculating the metrics, we sampled 100K points per mesh, and the threshold for computing the f-scores is set to 1cm.
To evaluate texture reconstruction, we render front and back-view images of the generated textured meshes using aitviewer~\cite{Kaufmann_Vechev_aitviewer_2022}.
During our evaluations, we noticed that some baselines, specifically PHORHUM~\cite{alldieck2022phorhum} and 2K2K~\cite{han2023high}, cannot handle non-front-facing images.
Therefore, in the manuscript (\cref{tab:3D}) all the results used front-facing images.
To provide a more comprehensive comparison and an evaluation aligned with real-world use cases, we include results based on images rendered from multiple view angles in~\cref{tab:3D_supp}. The CAPE and CustomHumans datasets contain images from three and four view angles respectively. Despite marginal degradation, the results indicate that our method consistently outperforms other methods in single-view 3D reconstructing.

\subsection{Robustness to View Angles}
\label{sec:angle_supp}

Inspired by insights from the previous subsection, we are interested in assessing the robustness of our method against variations in image view angles.
To this end, we rendered images by rotating the texture scans by $\{ 0, 15, 30, 45, 60, 75, 90\}$ degrees and subsequently computed their perspective 3D reconstruction metrics.
This analysis is detailed in~\cref{tab:angle_supp}. We found that our pipeline maintains robustness with viewpoint perturbations up to 45 degrees.
However, a significant increase in the Chamfer distance was observed when the angle increased from 45 to 60 degrees. 
This difference could stem from potential failures in pose estimation or the underlying assumption that human bodies can be reconstructed from only front and back-view images, which may not hold true at wider angles.
These observations provide a strong motivation for future research focused on enhancing the robustness of image reconstruction across varying view angles

\subsection{Verification of Design Choices}
\label{sec:design_supp}

\begin{figure*}[t]
\centering
\includegraphics[width=\linewidth]{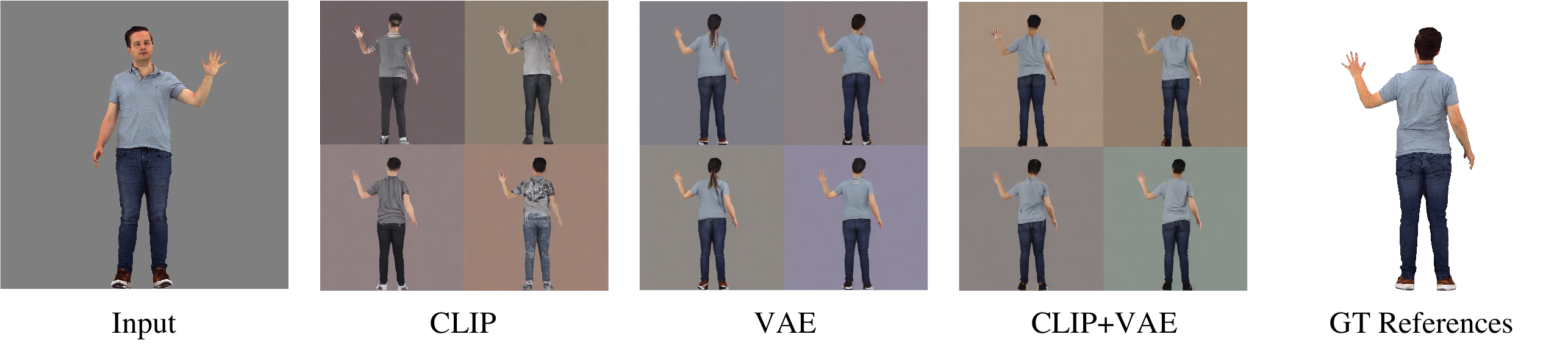}
\caption{\textbf{Analysis of different conditioning strategies}. We visualize the images generated by using different image features for conditioning. We show that combining both CLIP and VAE image features achieves more consistent and desirable results in back-view hallucination. Note that the four different images are sampled from different random seeds. Best viewed in color and zoom in.}
\label{fig:design_conditioning}
\vspace{-.5em}
\end{figure*}
\begin{figure*}[t]
\centering
\includegraphics[width=\linewidth]{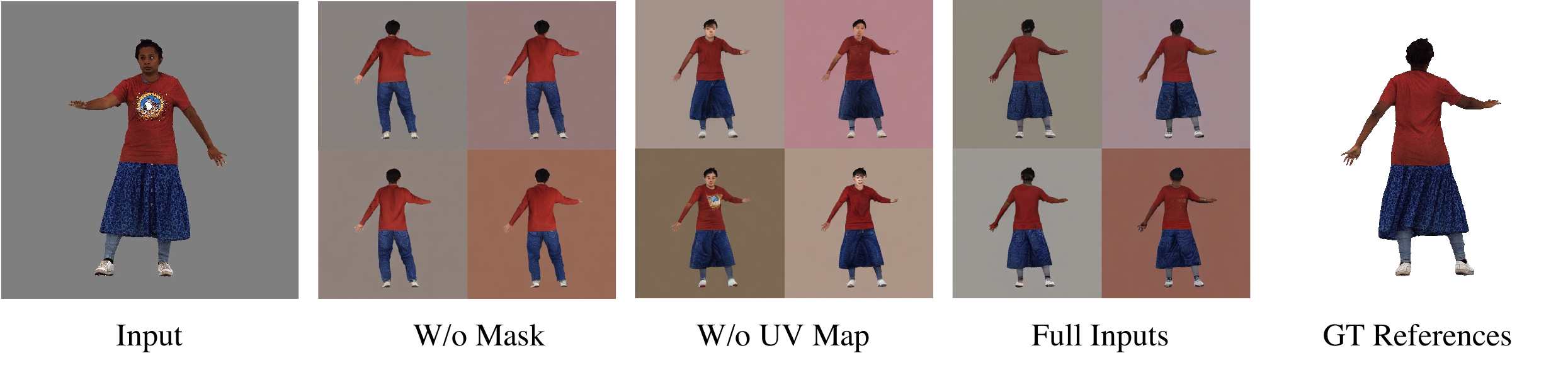}
\caption{\textbf{Analysis of different ControlNet conditions}. We visualize the images generated under various input conditions to the ControlNet model. We show that the integration of both UV maps and silhouette masks is crucial for generating spatially aligned back-view images. Note that the four different images are sampled from different random seeds. Best viewed in color and zoom in.}
\label{fig:design_input}
\vspace{-.5em}
\end{figure*}

\begin{figure*}[t]
\centering
\includegraphics[width=\linewidth]{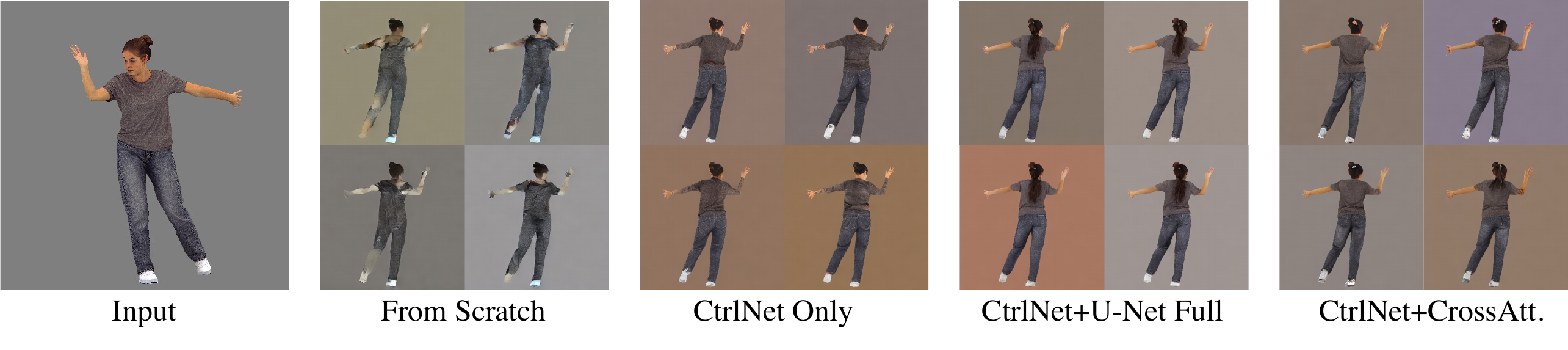}
\caption{\textbf{Analysis of different network training strategies}. We visualize the images generated by employing different network training strategies. We show that our method produces images with consistent appearances and is able to generate diverse hairstyles and clothing details. Note that the four different images are sampled from different random seeds. Best viewed in color and zoom in.}
\label{fig:design_finetune}
\vspace{-.5em}
\end{figure*}

\paragraph{Image conditioning strategies.}
We analyzed different strategies to incorporate image-conditioning in the diffusion U-Net.
~\cref{fig:design_conditioning} depicts the effects of using the CLIP image encoder and the VAE image encoder.
The results show that simply relying on the \textbf{CLIP} image encoder is not sufficient to provide accurate image conditioning. The clothing appearances cannot be accurately represented in the shared latent space of texts and images. On the other hand, the \textbf{VAE} encoder alone might also lose semantic information, such as male and female, for back-view hallucination. The hairstyles in the back are not consistent with the front-view image. Finally, the combination of both image features (\textbf{CLIP+VAE}) complements missing information of each image feature, therefore achieving more plausible results for back-view hallucination.

\paragraph{ControlNet inputs.}
We conducted controlled experiments to validate the efficacy of using SMPL-X UV maps and silhouette masks as conditioning inputs for our diffusion model.
~\cref{fig:design_input} illustrates the impact of employing different input images on the ControlNet models.
Our results show that omitting the silhouette masks (\textbf{w/o Mask}) results in output images that lack consistent body shapes with the input images, especially in areas with garments like skirts. 
Conversely, while relying solely on silhouette masks (\textbf{w/o UV Map}) ensures shape consistency, the model struggles to differentiate between front and back views.
This is particularly evident in the incorrect appearances on the head and face. 
Notably, the integration of both the silhouette and SMPL-X UV maps leads to more stable and accurate back-view hallucinations, thereby validating our approach.

\paragraph{Parameters finetuning.}
We conducted an analysis of the training strategy for our image-conditioned diffusion model by designing and comparing several training strategies.

\begin{itemize}
    \item We explored training from scratch, where both the diffusion U-Net and the ControlNet model are randomly initialized. This method is labeled as \textbf{From Scratch}.
    \item We initialized the model from a pretrained diffusion U-Net, but kept its parameters frozen during training. In this variant, only the ControlNet model's parameters are optimized, and it is denoted as \textbf{CtrlNet Only}.
    \item We developed a strategy that unfreezes all parameters in both the pretrained diffusion U-Net and the ControlNet model for training, referred to as \textbf{CtrlNet+U-Net Full}.
    \item Finally, we presented the training strategy used in our method, i.e., training only the cross-attention layers in the pretrained diffusion U-Net along with the ControlNet model (denoted as \textbf{CtrlNet+CrossAtt.}).
\end{itemize}

The results of these training strategies are depicted in~\cref{fig:design_finetune}, which shows that training a large diffusion model from scratch using only 500 3D scans is impractical. 
While leveraging large-scale pretraining can mitigate this issue, the CtrlNet Only training strategy fails to generate consistent appearances from front-view images.
Alternatively, when we unfroze the parameters in the diffusion U-Net, the model showed improvement in generating images more aligned with the input conditional image. 
However, this approach led to a limitation where the model consistently produced identical output images, thus compromising its generative capability, particularly in varying clothing wrinkles and hairstyles. 
In contrast, our training strategy successfully generates perceptually consistent back-view images while preserving the model’s generative capabilities. This strategy effectively handles the stochastic nature of clothing details and hairstyles for back-view hallucination.
\begin{table}[t]
    \centering
    \small
\begin{tabular}{lccc}
\toprule
 Method               & \begin{tabular}[c]{@{}c@{}}Pred-to-Scan /\\ Scan-to-Pred (mm)$\downarrow$ \end{tabular} & NC$\uparrow$ & f-Score$\uparrow$     \\
\midrule
W/o Normal     & \textbf{16.825} / \textbf{19.802} &\textbf{ 0.837} & \textbf{40.593}   \\
W/ Normal      & 18.709 / 20.451 & 0.826  & 37.029   \\
\bottomrule
\end{tabular}
\caption{\textbf{Effectiveness of normal guidance}. We verify the effectiveness of incorporating normal guidance in our pipeline. While we observed that normal guidance marginally reduces performance in terms of 3D metrics, it significantly enhances the overall perceptual quality.}
\vspace{-1.5em}

\label{tab:normal}
\end{table}

\begin{figure}[t]
\centering
\includegraphics[width=\linewidth]{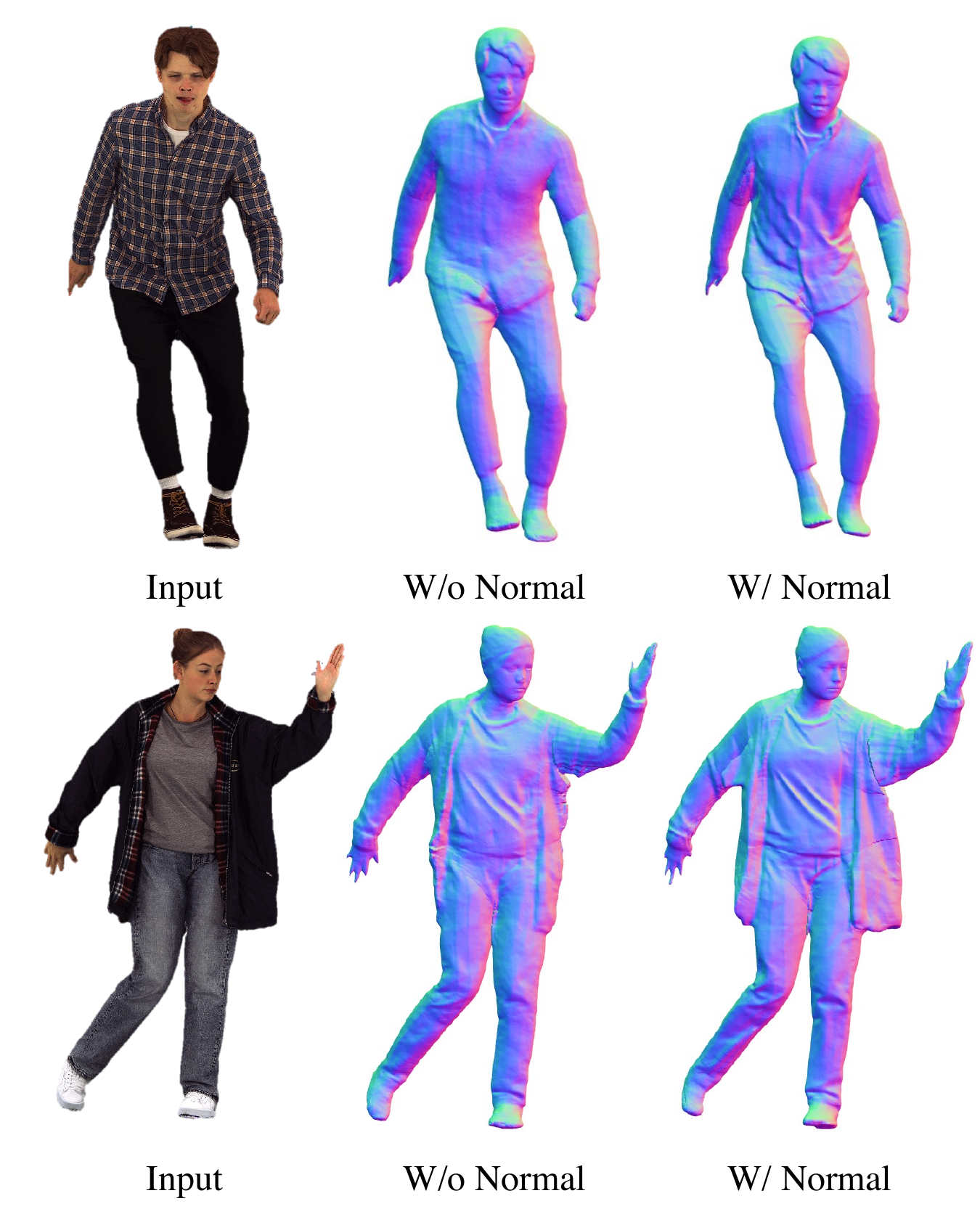}
\caption{\textbf{Visualization of the effectiveness of normal guidance}. The use of normal guidance demonstrates its superiority in capturing geometric details of clothing and is more robust in reconstructing a challenging coat. }
\label{fig:design_normal}
\vspace{-1em}
\end{figure}

\paragraph{Importance of normal guidance.}
We validated the use of normal guidance in our mesh reconstruction module.
In this experiment, we created a variant where normal images were replaced with RGB images during local feature querying. 
The 3D reconstruction results, as shown in~\cref{tab:normal}, surprisingly indicate that this variant surpasses our model with normal guidance across all metrics.
However, \cref{fig:design_normal} illustrates the tangible benefits of incorporating normal guidance. 
Without normal guidance, the mesh surface becomes noticeably smoother, and the model struggles to accurately reconstruct challenging clothing, such as coats. 
This observation aligns with our findings in~\cref{sec:ablation} and~\cref{sec:plugin_supp}, indicating that conventional 3D metrics may not fully capture perceptual quality. 
Hence, this trade-off highlights the importance of the normal predictor and guidance in achieving high-fidelity 3D human reconstruction.

\subsection{Additional Results}
\begin{figure*}[t]
\centering
\includegraphics[width=\linewidth]{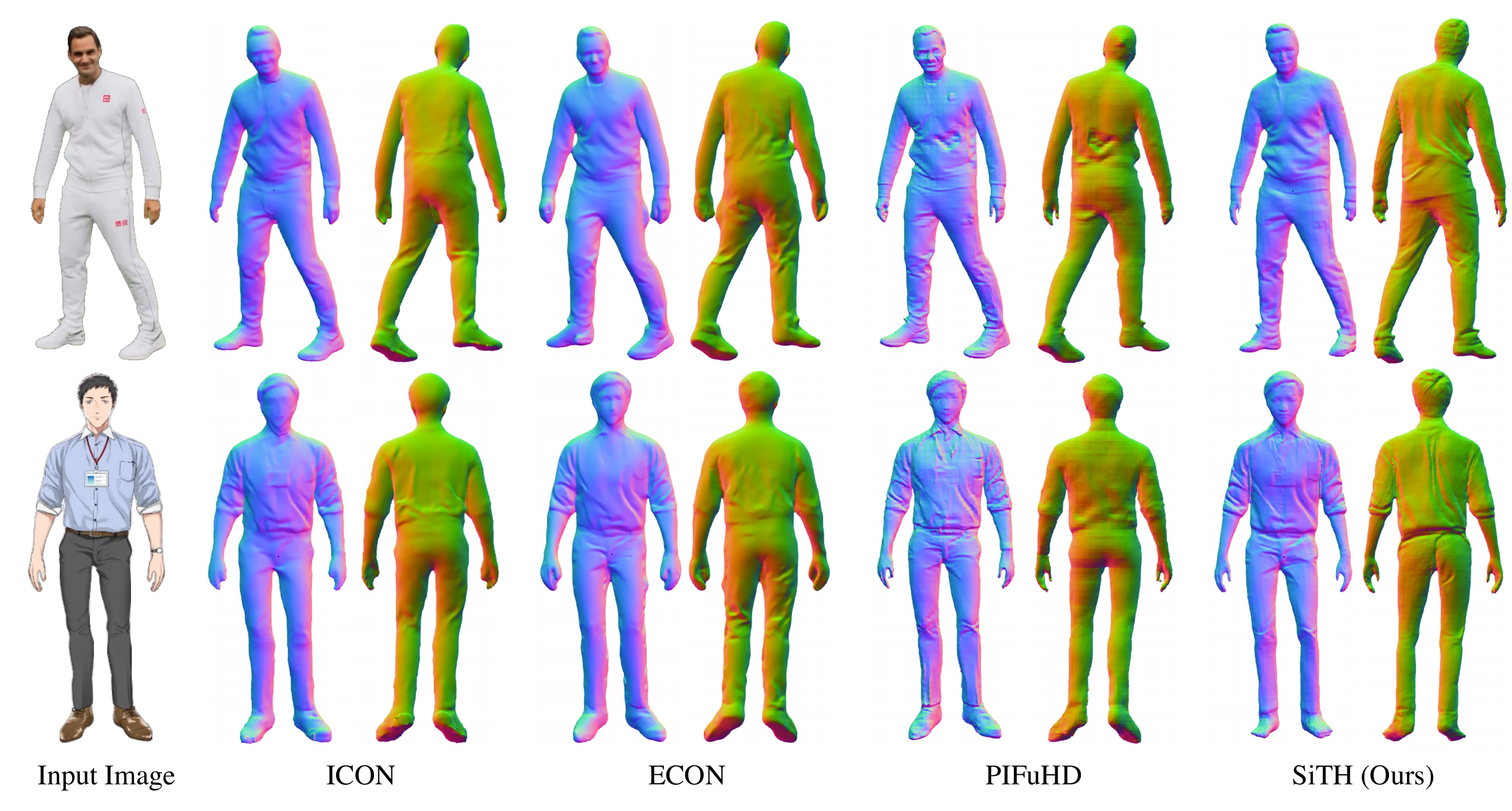}
\vspace{-2em}
\caption{\textbf{Qualitative comparison of mesh geometry with Internet images}. Our method generates realistic clothing wrinkles in the back regions. Best viewed in color and zoom in.}
\label{fig:3D_geo_supp}
\vspace{-.5em}
\end{figure*}

\begin{figure*}[t]
\centering
\includegraphics[width=\linewidth]{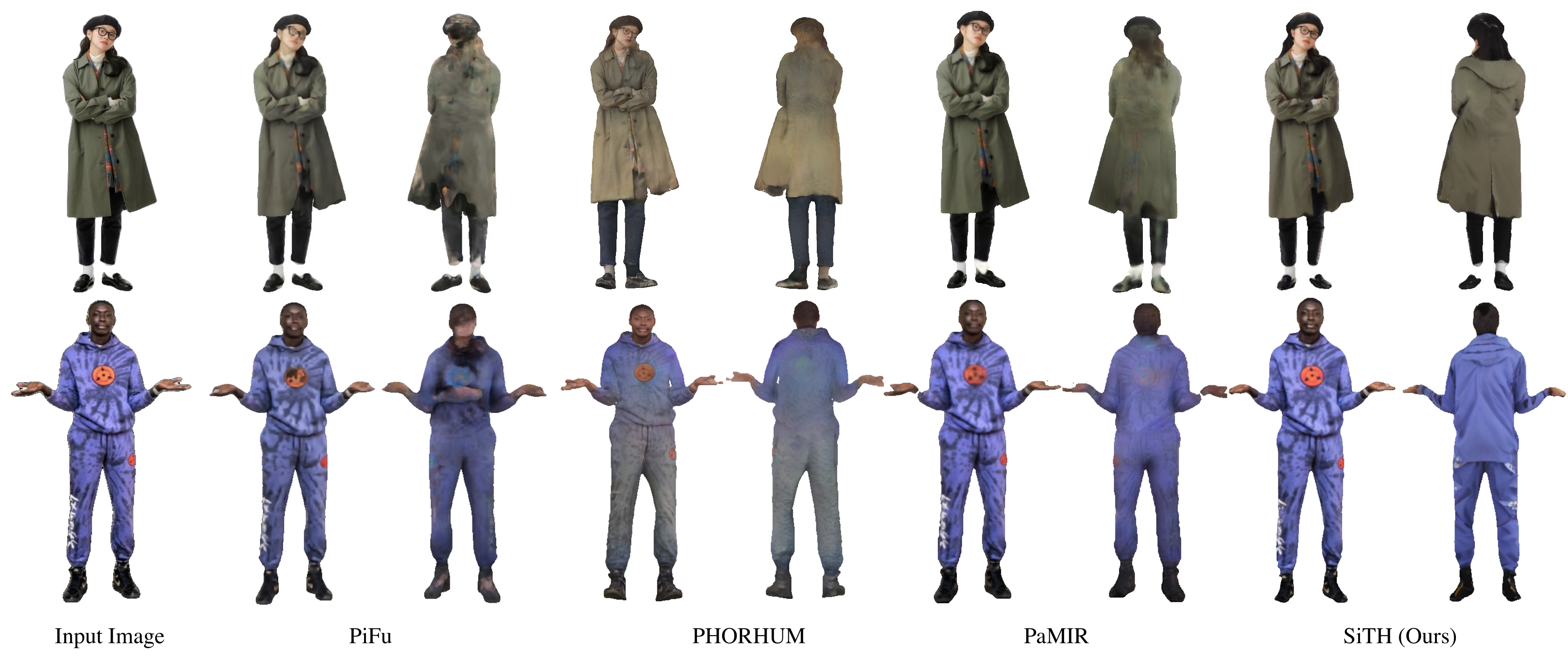}
\vspace{-2em}
\caption{\textbf{Qualitative comparison of mesh texture with Internet images}. Our method generates realistic texture in and back regions. Best viewed in color and zoom in.}
\label{fig:3D_tex_supp}
\vspace{-.5em}
\end{figure*}

\begin{figure*}[t]
\centering
\includegraphics[width=\linewidth]{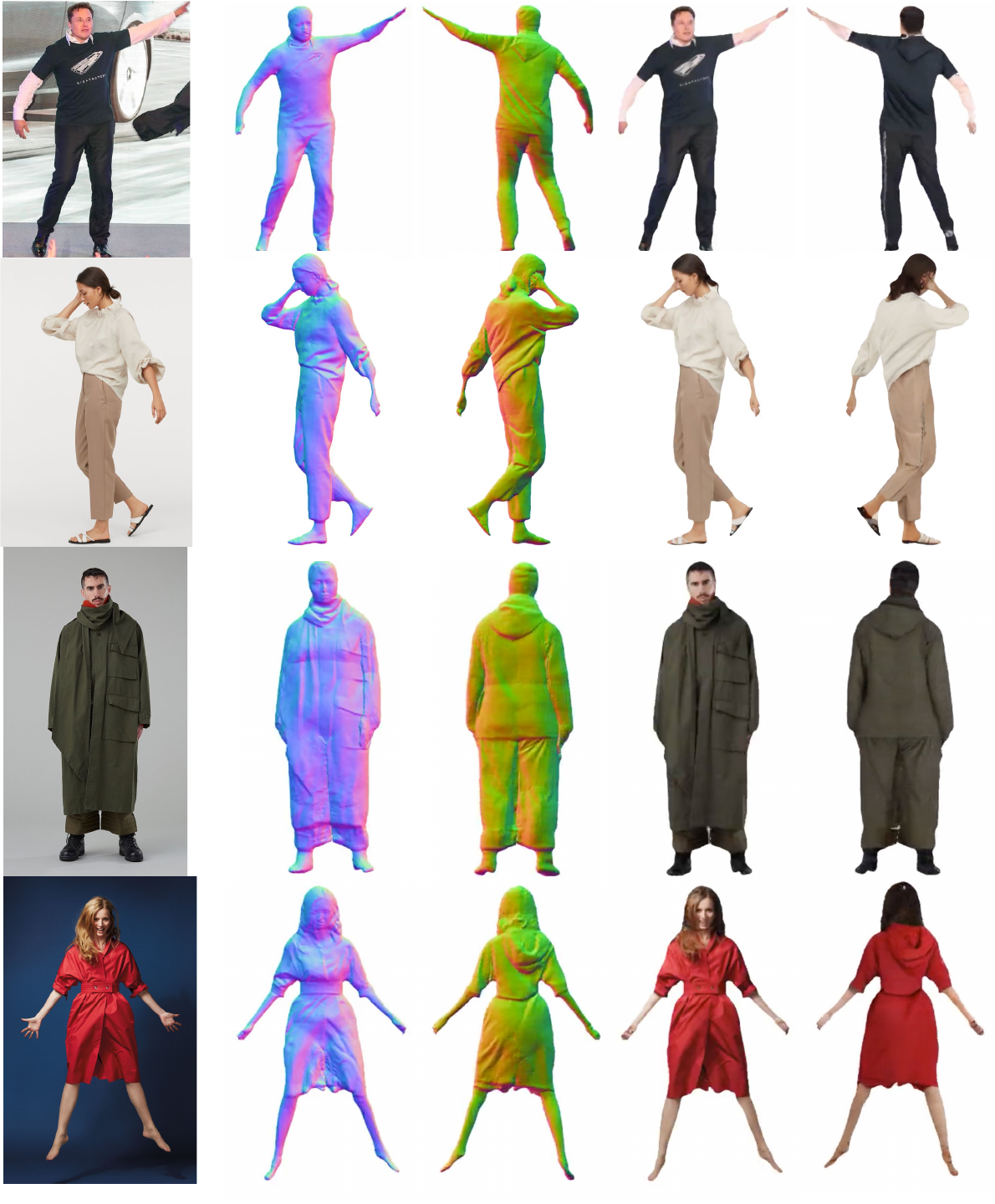}
\vspace{-2em}
\caption{\textbf{Examples of reconstruction from Internet images}. Our method generates realistic clothing wrinkles in the back regions. Best viewed in color and zoom in.}
\label{fig:large_supp}
\vspace{-.5em}
\end{figure*}

\begin{figure*}[t]
\centering
\includegraphics[width=\linewidth]{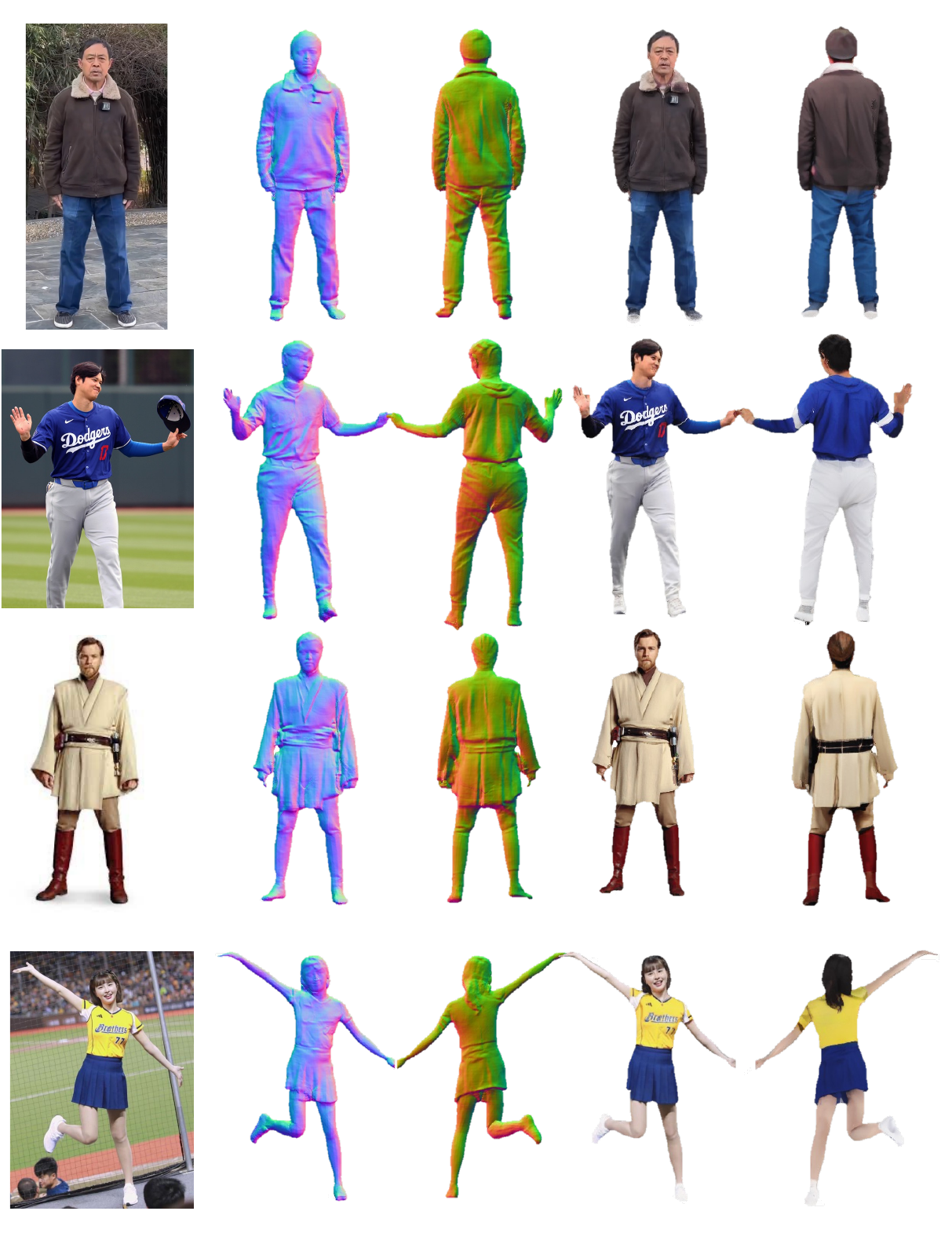}
\vspace{-2em}
\caption{\textbf{Examples of reconstruction from Internet images}. Our method generates realistic clothing wrinkles in the back regions. Best viewed in color and zoom in.}
\label{fig:large2_supp}
\vspace{-.5em}
\end{figure*}

We present more qualitative results in~\cref{fig:3D_geo_supp}, \cref{fig:3D_tex_supp}, \cref{fig:large_supp}, and~\cref{fig:large2_supp}, demonstrating our method's robustness in handling unseen images sourced from the Internet.

\section{Discussion}
\begin{figure}[t]
\centering
\includegraphics[width=\linewidth]{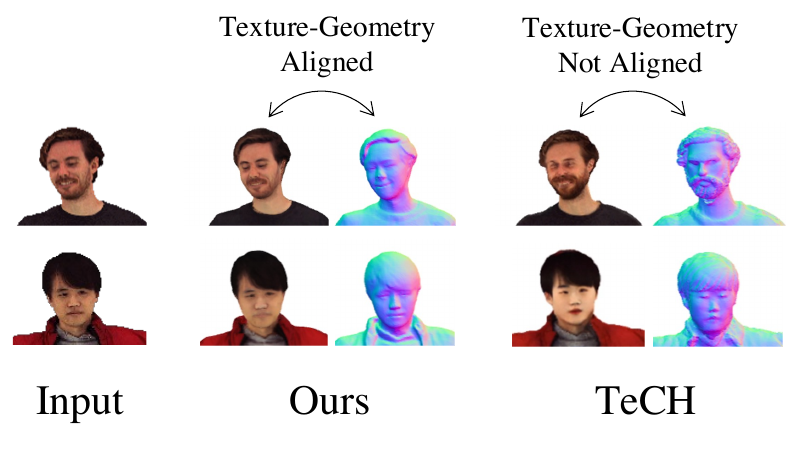}
\caption{\textbf{Comparison with SDS optimization-based method}. Compared to the optimization-based method (TeCH), our method reconstructs consistent facial details and well-aligned mesh texture and geometry. Note that TeCH requires 6 hours to optimize both texture and geometry.}
\vspace{-1em}
\label{fig:vs_tech}
\end{figure}

\subsection{Data-driven v.s. Optimization} 
\label{sec:discuss_sds}
Numerous concurrent works, such as TECH~\cite{huang2024tech} and Human-SGD~\cite{albahar2023humansgd}, propose creating 3D textured humans from single images using optimization-based approaches.
These methods primarily build upon pretrained diffusion models and a Score Distillation Sampling loss, with several adaptations. 
In our discussion, we highlight the unique aspects of our method in comparison.
Our method uniquely integrates a diffusion model into the existing data-driven 3D reconstruction workflow. 
This integration allows us to efficiently exploit 3D supervision to learn a generalized model for single-view reconstruction, thus avoiding the need for costly and time-consuming per-subject optimization.
Consequently, our pipeline can generate high-quality textured meshes in under two minutes.
Moreover, we observed that the existing optimization-based methods failed to generate 3D meshes having consistent and aligned texture and geometry (\cref{fig:vs_tech}). This is due to their requirements of optimizing texture and geometry with separate optimization processes. Instead, our results are more similar to the input images and retain the consistency of both texture and geometry.
Lastly, our two-stage pipeline empowers the 3D human creation process with controllability.
As demonstrated in~\cref{sec:compare2D_supp}, our hallucination model handles generative stochasticity and is able to create various plausible back-view images.
This feature provides users with the flexibility to choose back-view appearances based on their preferences, instead of solely relying on a random optimization process.
However, as previously discussed, our method does have certain limitations.
We believe that the further cross-pollination of both methods offers a promising path for future developments in generative 3D human creation.

\begin{figure}[t]
\centering
\includegraphics[width=\linewidth]{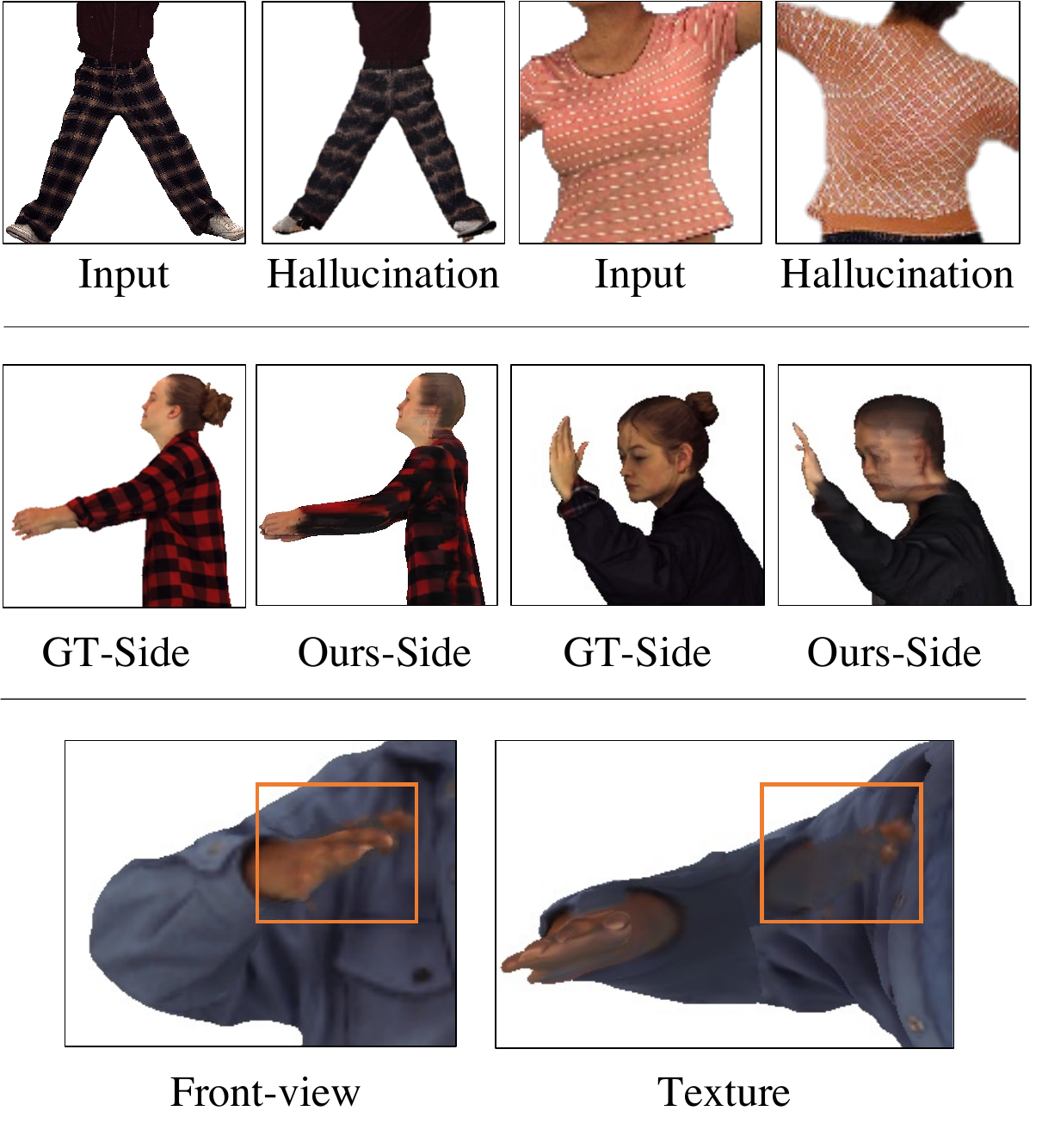}
\caption{\textbf{Limitations}. \emph{Top}: The hallucination model struggles with complex textures like stripes or plaid. \emph{Middle}: Side-view appearances are not accurately recovered by mesh reconstruction. \emph{Bottom}: The mesh reconstruction model is unable to effectively handle self-occluded regions.}
\vspace{-1em}
\label{fig:fail}
\end{figure}

\subsection{Limitations}

\paragraph{Complex clothing textures.}
We observed a challenge with the image-conditioned diffusion model in accurately generating complex clothing textures, such as stripes or plaid (\cref{fig:fail}~\emph{Top}).
This limitation stems from the image feature resolution using a pretrained VAE image encoder for feature extraction and reconstruction.
The model generates output images at a resolution of $512 \times 512$, yet the diffusion U-Net is limited to processing features of only $64 \times 64$.
Consequently, finer texture details may be lost in the diffusion process.
This issue motivates the need for future development of pixel-perfect image-conditioning approaches, which could more accurately capture details in high-resolution images.

\paragraph{Side-view appearances.}
Our method follows the established practice in single-view human reconstruction, using a "sandwich-like" approach that relies on front and back information.
This technique reduces the need for extensive multi-view images for 3D reconstruction. However, as shown in~\cref{fig:fail}~\emph{Middle}, a limitation of this method is the loss of detail in side views.
A promising direction for future enhancements would be integrating our pipeline with optimization-based methods for a more detailed 3D human creation.
Our pipeline currently provides a robust initialization by providing 3D human models with geometric and appearance details. 
By leveraging this initialization, the lengthy optimization process could be accelerated, making it more effective for creating detailed 3D humans.

\paragraph{Self-occlusion.}
Our mesh reconstruction module struggles to reconstruct appearance details in self-occluded regions, as illustrated in~\cref{fig:fail}~\emph{Bottom}. 
This challenge arises because essential information in these areas is not captured by either front or back-view images, and thus the mesh reconstruction module fails to infer these details.
One potential solution is using an optimization process for refinement, as previously suggested. 
Another promising direction for future work could be developing a hallucination model capable of generating multi-view images with accurate 3D consistency, which would help reduce the self-occluded regions in mesh reconstruction.



\end{document}